%% file: main.tex
\documentclass[12pt]{article}

\usepackage{times}
\usepackage{xcolor}
\usepackage{soul}
\usepackage[utf8]{inputenc}
\usepackage[small]{caption}
\usepackage{xspace}

\usepackage{graphicx}
\usepackage{collcell}

\usepackage{array,multirow}
\usepackage{booktabs}

\usepackage[all,knot]{xy}
\usepackage{amsmath}
\usepackage{amssymb}

\usepackage[hidelinks]{hyperref}

\input{abbreviations-v3}
\usepackage{xspace}

\usepackage{cleveref}

\usepackage{calc}
\usepackage{amssymb}
\usepackage{amstext}
\usepackage{amsmath}

\usepackage {tikz}
\usepackage{mathtools}
\usepackage{tikz-cd}
\usetikzlibrary{decorations.pathmorphing}
\usetikzlibrary{arrows}
\usetikzlibrary{arrows.meta,calc}
\usetikzlibrary {positioning}
\usetikzlibrary{shadows}
\tikzset{
diagonal fill/.style 2 args={fill=#2, path picture={
\fill[#1, sharp corners] (path picture bounding box.south west) -|
                         (path picture bounding box.north east) -- cycle;}},
reversed diagonal fill/.style 2 args={fill=#2, path picture={
\fill[#1, sharp corners] (path picture bounding box.north west) |- 
                         (path picture bounding box.south east) -- cycle;}}
}

\usepackage[margin=1in]{geometry}
\usepackage{apacite}
    
\title{Cognitive Argumentation and the Suppression Task}

\author{Emmanuelle-Anna Dietz Saldanha,$^\text{a}$ Antonis Kakas$^\text{b}\thanks{The authors are mentioned in alphabetical order.}$ \quad\quad\bigskip \\
\small $^a$ International Center for Computational Logic, TU
  Dresden, Germany, emmanuelle.dietz@tu-dresden.de \\
\small  $^b$ 
  Department of Computer Science, University of Cyprus, Nicosia, Cyprus, antonis@ucy.ac.cy
  }
\date{} 

\begin{document}

\maketitle

\begin{abstract}

    This paper addresses the challenge of modeling human reasoning, within 
    a new framework called Cognitive Argumentation.  
    This framework rests on the assumption that human logical reasoning is inherently a process of 
    dialectic argumentation and aims to develop a cognitive model 
    for human reasoning that is computational and implementable. 
    To give logical reasoning a human cognitive form the framework 
    relies on cognitive principles, based on
    empirical and theoretical work in Cognitive Science, to suitably adapt a
    general and abstract framework of computational argumentation from AI. 

    The approach of Cognitive Argumentation is evaluated with respect to 
    Byrne's suppression task, where the aim is not only to capture the 
    suppression effect between different groups of people but also to 
    account for the variation of reasoning within each group. 
    Two main cognitive principles are particularly important to 
    capture human conditional reasoning that explain the participants' responses: 
    (i) the interpretation of a condition within a conditional as sufficient and/or necessary and
    (ii) the mode of reasoning either as predictive or explanatory. 
    We argue that Cognitive Argumentation provides a coherent and cognitively adequate model for human conditional reasoning that allows a natural distinction between definite and plausible conclusions, exhibiting the important characteristics of context-sensitive and defeasible reasoning.\end{abstract}

\section{Introduction}

\input{introduction-short.tex}

\section{The Suppression Task} \label{sect:informal}

\input{informal.tex}

\section{Cognitive Principles} \label{sect:cogprin}

\input{cognitiveprinciples-short.tex}

\section{Cognitive Argumentation} \label{sect:argumentation}

\input{argumentation.tex}

\section{The Suppression Task} \label{sect:bst}

\input{bst-short.tex}

\section{\COGNICA \ -- Cognitive Argumentation on the Web} \label{sect:cognica}

\input{cognica.tex}

\section{Discussion} \label{sect:discussion}

\input{discussion-short.tex}

\section{Conclusion}

\input{conclusions.tex}

\bibliographystyle{apacite-no-initials}
\bibliography{bib} 

\end{document}

%% file: abbreviations-v3.tex
\definecolor{mylightgrayB}{RGB}{213,222,229}
\definecolor{mylightgrayC}{RGB}{233,252,255}

\newcommand{\mycolorG}{mylightgrayC!80!gray}

\newcommand{\mycolorR}{white}

\DeclareSymbolFont{lasy}{U}{lasy}{m}{n}
\SetSymbolFont{lasy}{bold}{U}{lasy}{b}{n}
\DeclareMathSymbol\leadsto{\mathrel}{lasy}{"3B}
\newcommand{\imply}{\leadsto}

\definecolor{light-gray}{gray}{0.90}
\newcommand{\high}{\colorbox{black!25}}
\newcommand{\lesshigh}{\colorbox{black!10}}

\newcommand{\borderlesshigh}{\fcolorbox{black}{black!10}}

\newcommand{\COGNICA}{\textsf{COGNICA}\xspace}

\newcommand{\condition}{\textit{condition}\xspace}

\newcommand{\consequence}{\textit{consequence}\xspace}
\newcommand{\suf}{\small\textsf{suff}\xspace}
\newcommand{\nec}{\small\textsf{necc}\xspace}

\newcommand{\consq}{\mbox{consq}\xspace}
\newcommand{\cond}{\mbox{cond}\xspace}

\newcommand{\nconsq}{\overline{\mbox{consq}}\xspace}
\newcommand{\ncond}{\overline{\mbox{cond}}\xspace}

\newcommand{\defname}[1]{\textit{#1}}

\makeatletter
\newcolumntype{G}{>{\collectcell\@gobble}c<{\endcollectcell}@{}}
\makeatother

{\bfseries}{\rmfamily}

\newcommand{\needmilk}{\varneedm \imply \buymilk}
\newcommand{\asksmilk}{\varasksm \imply \buymilk}
\newcommand{\moneymilk}{\varmoneym \imply \buymilk}

\newcommand{\didntbuy}{\overline{\mathit{buy}}}
\newcommand{\buymilk}{\varbuym}

\newcommand{\varmilk}{\mathit{milk}}
\newcommand{\varneedm}{\mathit{need}}
\newcommand{\varbuym}{\mathit{buy}}
\newcommand{\varasksm}{\mathit{asks}}
\newcommand{\varmoneym}{\mathit{money}}

\newcommand{\nvarneedm}{\overline{\varneedm}}
\newcommand{\nvarbuym}{\overline{\varbuym}}
\newcommand{\nvarasksm}{\overline{\varasksm}}
\newcommand{\nvarmoneym}{\overline{\varmoneym}}

\newcommand{\nmoneymilk}{\nvarmoneym \imply \nvarbuym}

\newcommand{\buyneed}{\varbuym \imply \varneedm}

\newcommand{\nbuymoney}{\nvarbuym \imply \nvarmoneym}

\newcommand{\needDneedmilk}{\Delta^\varneedm_{\varneedm \overset{\textit{}}{\imply} \buymilk}}

\newcommand{\needDneedmilkmoney}{\Delta^\varneedm_{\varneedm \overset{\textit{}}{\imply} \buymilk, m}}

\newcommand{\nmonyDmoneymilk}{\Delta^{\nvarmoneym}_{\nvarmoneym \overset{\textit{}}{\imply}  \nvarbuym}}

\newcommand{\Dnmoneymilk}{\Delta_{\nvarmoneym \overset{\textit{n}}{\imply}  \nvarbuym}}

\newcommand{\lD}{\Delta^\lib}

\newcommand{\eD}{\Delta^{e}}

\newcommand{\Dl}{\Delta_\lib}
\newcommand{\Do}{\Delta_\open}

\newcommand{\De}{\Delta_e}

\newcommand{\Dne}{\Delta_{\ness}}

\newcommand{\Dno}{\Delta_{\nopen}}
\newcommand{\Dnl}{\Delta_{\nlib}}

\newcommand{\Del}{\Delta_{e, e \overset{\textit{s}}{\imply} \lib}}
\newcommand{\eDel}{\Delta_{e \overset{\textit{s}}{\imply} \lib}^{\mathit{e}}}

\newcommand{\nlDol}{\Delta_{\nlib  \overset{\textit{n}}{\imply} \overline{o}}^{\overline{\lib}}}

\newcommand{\Dtl}{\Delta_{{t \overset{\textit{s}}{\imply} \lib}}}

\newcommand{\Dcondle}{\Delta_{\ell  \overset{\textit{n}}{\imply}  e}}
\newcommand{\DcondleSuff}{\Delta_{\ell \overset{\textit{s}}{\imply} e}}
\newcommand{\lDle}{\Dcondle^{\ell}}
\newcommand{\lDleSuff}{\DcondleSuff^{\ell}}

\newcommand{\Dcondnlne}{\Delta_{\overline{\ell} \overset{\textit{s}}{\imply} \overline{e}}}
\newcommand{\nlDcondnlne}{\Dcondnlne^{\nlib}}

\newcommand{\DcondnlnexoN}{\Delta_{\overline{\ell} \overset{\textit{n}}{\imply} \overline{\textsf{\tiny exo}}}}

\newcommand{\DcondnlneN}{\Delta_{\overline{\ell} \overset{\textit{n}}{\imply} \overline{e}}}
\newcommand{\DcondnlneS}{\Delta_{\overline{\ell} \overset{\textit{s}}{\imply} \overline{e}}}
\newcommand{\DcondnlntS}{\Delta_{\overline{\ell} \overset{\textit{s}}{\imply} \overline{t}}}
\newcommand{\nlDcondnlneN}{\DcondnlneN^{\nlib}}
\newcommand{\nlDcondnlneS}{\DcondnlneS^{\nlib}}
\newcommand{\nlDcondnlntS}{\DcondnlntS^{\nlib}}

\newcommand{\Dnel}{\Delta_{\ness, \ness \overset{\textit{n}}{\imply} \nlib}}

\newcommand{\Dnol}{\Delta_{\overline{o}, \overline{o} \overset{\textit{n}}{\imply} \nlib}}

\newcommand{\neD}{\Delta^{\overline{\mathit{e}}}}
\newcommand{\neDel}{\Delta_{\ness  \overset{\textit{n}}{\imply} \nlib}^{\overline{\mathit{e}}}}
\newcommand{\neDnel}{\Delta_{\ness  \overset{\textit{n}}{\imply} \nlib}^{\ness}}

\newcommand{\essay}{\mathit{e}}
\newcommand{\nessay}{\mathit{\overline{e}}}
\newcommand{\library}{\lib}
\newcommand{\nlibrary}{\nlib}
\newcommand{\lib}{\ell}
\newcommand{\nlib}{\overline{\ell}}

\newcommand{\ness}{\overline{e}}
\newcommand{\tbook}{t}
\newcommand{\ntbook}{\overline{t}}
\newcommand{\open}{o}
\newcommand{\nopen}{\overline{o}}

\newcommand{\smpl}{e \imply \lib}
\newcommand{\altn}{\tbook \imply \lib}
\newcommand{\add}{\open \imply \lib}
\newcommand{\bialtn}{\lib \imply t}
\newcommand{\bismpl}{\lib \imply e}
\newcommand{\bialt}{\lib \imply \tbook}
\newcommand{\biadd}{\lib \imply \open}
\newcommand{\negadd}{\overline{\open} \imply \overline{\lib}}
\newcommand{\negsmpl}{\overline{e} \imply \overline{\lib}}

\newcommand{\negbiadd}{\overline{\lib} \imply \overline{\open}}
\newcommand{\negbismpl}{\overline{\lib} \imply\overline{e}}
\newcommand{\negbialt}{\overline{\lib} \imply \overline{\tbook}}

\newcommand{\CalL}{\mathcal{L}}
\newcommand{\CalA}{\mathcal{A}}
\newcommand{\CalB}{\mathcal{B}}

\newcommand{\CalS}{\mathcal{S}}

\newcommand{\CalC}{\mathcal{C}}
\newcommand{\CalP}{\mathcal{P}}

\newcommand{\Bst}{{\CalB\textsf{\tiny st}}}

\newcommand{\ALF}{\CalA_\CalL}
\newcommand{\ALB}{\CalA_\Bst}

\newcommand{\ASset}{\CalA s}
\newcommand{\Cf}{\CalC}
\newcommand{\St}{\succ}

\newcommand{\Prem}{\CalS}
\newcommand{\Pre}{\small\textsf{pre}}
\newcommand{\Pos}{\small\textsf{pos}}

\newcommand{\suffexo}{\mbox{exo\_e}}
\newcommand{\exo}{\small\mbox{exo}}

\newcommand{\lDexo}{\Delta^\ell_{\ell \overset{\textit{s}}{\imply} \textsf{\tiny exo}}}

\newcommand{\lDlt}{\Delta^\ell_{\ell \overset{\textit{s}}{\imply} \textsf{\tiny t}}}

\newcommand{\AS}{\small\textsf{as}}

\newcommand{\hyp}[1]{{\textsf{\small hyp}}(#1)}

\newcommand{\ehyp}{{\textsf{\small hyp}}}
\newcommand{\thyp}[1]{{\textsf{\tiny hyp}}(#1)}

\newcommand{\eFact}{{\textsf{\small fact}}}
\newcommand{\Fact}[1]{{\textsf{\small fact}}(#1)}

\newcommand{\cogPrinFact}{{\small\textsf{maxim of quality}}}
\newcommand{\cogPrinHyp}{{\small\textsf{maxim of relevance}}}

\newcommand{\psuff}{\mbox{suff\_p}}
\newcommand{\pnecc}{\mbox{necc\_p}}
\newcommand{\enecc}{\mbox{necc\_e}}
\newcommand{\esuff}{\mbox{suff\_e}}
\newcommand{\secpsuff}{\mbox{sec\_suff\_p}}
\newcommand{\secpnecc}{\mbox{sec\_necc\_p}}
\newcommand{\secesuff}{\mbox{sec\_suff\_e}}

\newcommand{\eASpsuff}{{\textsf{\small suff\_p}}}
\newcommand{\eASpnecc}{{\textsf{\small necc\_p}}}
\newcommand{\eASenecc}{{\textsf{\small necc\_e}}}
\newcommand{\eASesuff}{{\textsf{\small suff\_e}}}
\newcommand{\eASsecpsuff}{{\textsf{\small sec\_suff\_p}}}
\newcommand{\eASsecesuff}{{\textsf{\small sec\_suff\_e}}}
\newcommand{\eASsecpnecc}{{\textsf{\small sec\_necc\_p}}}

\newcommand{\ASpsuff}[1]{{\textsf{\small suff\_p}}(#1)}
\newcommand{\ASpnecc}[1]{{\textsf{\small necc\_p}}(#1)}
\newcommand{\ASenecc}[1]{{\textsf{\small necc\_e}}(#1)}
\newcommand{\ASesuff}[1]{{\textsf{\small suff\_e}}(#1)}
\newcommand{\ASsecpsuff}[1]{{\textsf{\small sec\_suff\_p}}(#1)}
\newcommand{\ASsecpnecc}[1]{{\textsf{\small sec\_necc\_p}}(#1)}

\newcommand{\ASsecesuff}[1]{\secesuff(#1)}

\newcommand{\cogPrinPsuff}{{\small\textsf{sufficient prediction}}\xspace}
\newcommand{\cogPrinPnecc}{{\small\textsf{necessary prediction}}\xspace}
\newcommand{\cogPrinsecPsuff}{{\small\textsf{secondary sufficient prediction}}\xspace}
\newcommand{\cogPrinsecEsuff}{{\small\textsf{secondary sufficient explanation}}\xspace}
\newcommand{\cogPrinsecPnecc}{{\small\textsf{secondary neccessary prediction}}\xspace}
\newcommand{\cogPrinEsuff}{{\small\textsf{sufficient explanation}}\xspace}
\newcommand{\cogPrinEnecc}{{\small\textsf{necessary explanation}}\xspace}
\newcommand{\cogPrinsecEsuffabbr}{{\small\textsf{sec. sufficient explanation}}\xspace}
\newcommand{\cogPrinsecPsuffabbr}{{\small\textsf{sec. sufficient prediction}}\xspace}
\newcommand{\cogPrinsecPneccabbr}{{\small\textsf{sec. neccessary prediction}}\xspace}

\newcommand{\cogPrincExo}{{\small\textsf{exogenous explanation}}\xspace}

\newcommand{\explAS}{\mbox{e}}
\newcommand{\expl}{\mbox{expl}}
\newcommand{\obs}{\mbox{obs}}

\newcommand{\FPrem}{\mathcal F}
\newcommand{\HPrem}{\mathcal A}

%% file: introduction-short.tex
How do humans reason? What conclusions do they draw?
Can we provide a satisfactory explanation to these questions by means of a coherent,
computational and cognitively adequate model? 
A computational model that is able to reproduce human reasoning in a faithful way while at the same time
accounting for the variability in the reasoning observed across the population?

In this paper we aim to contribute in addressing this challenge
by formulating and studying human reasoning within the framework of Cognitive Argumentation.
This is a framework that is built by synthesizing together the general and abstract theory of computational argumentation from AI
with cognitive principles born out of empirical and theoretical findings of Cognitive Science.
We will examine how Cognitive Argumentation could give a new underlying formal basis for human reasoning
through an in depth and extensive analysis of Byrne's~\citeyear{byrne:89} suppression task, one of the most well-known psychological experiments on human (conditional) reasoning.

\emph{But why argumentation?}
Humans reason with knowledge whose generic form is that of an association or link between different
pieces of information. In contrast to formal logical reasoning which is strict and rigid and rarely matches that of 
human reasoning at large, argumentation provides a more flexible logical framework,
both in the representation of knowledge, that is
closer to the generic form of associations and in
the actual process of reasoning to conclusions. It is a
framework well suited for handling conflicting and dynamically changing
information, as indeed is the case we are confronted with in our
everyday human-level reasoning.

Support for argumentation exists from the early work of Aristotle
on dialectic syllogistic reasoning to numerous works in Cognitive Science and Philosophy
in the last century and more recently in AI. Two recent results
provide direct support in favour of the alternative of argumentation
for human reasoning. Firstly, there is now strong evidence from Cognitive Psychology in various studies,
brought together in the work of Mercier and Sperber~\citeyear{mercier:sperber:2011},
that humans arrive at conclusions and justify these by arguments.
Arguments are the means for human reasoning and the process of reasoning is
one of evaluation, elaboration and acceptance or rejection of arguments.
Secondly, recent results have shown that such an argumentative form
of reasoning, or Argumentation Logic as it is called in \cite{ALStudia2018,Informalizing},
can be arranged to give, as a special case, a reasoning process that is
completely equivalent to classical logical entailment.
Hence the departure of argumentation from formal logic is not radical, but instead one that 
uniformly encompasses both formal and informal reasoning.

\subsection{Cognitive Argumentation}

Cognitive Argumentation emerges out of the synthesis of formal argumentation theory in AI and
empirical and theoretical studies of the psychology of reasoning from
Cognitive Psychology and Philosophy. Formal argumentation in AI provides
a good computational basis for argumentation but for this to become a
cognitive model for human reasoning it needs to be informed and
guided by \emph{cognitive principles} of human thinking that have been emerging
out of studies in Cognitive Psychology over many decades now.
Cognitive Argumentation therefore puts an emphasis on accommodating empirical observations
of human reasoning and letting this phenomenology guide its development.

In realizing Cognitive Argumentation we need to consider
how cognitive principles affect and to a certain extent determine the
construction and evaluation of arguments. The construction of arguments would be
based on cognitive argument schemes (a central notion in the study of argumentation 
\cite{Toumlin1958,pollock:1995,walton:1996}) that capture the typical common sense
knowledge about our physical world or about our human behavior,
physical, mental or emotional. 
%
Cognitive principles should also guide the selection of which cognitively valid argument
schemes to actually use under the different dynamically changing conditions of the environment in which
the argumentation reasoning process takes place. This selection depends on a process of
\textit{awareness} of relevant argument schemes under the current circumstances
which in turn is guided by \textit{belief biases} and other \textit{extra-logical assumptions}
made by humans.

The dynamic nature of real-life human reasoning presents a major challenge for
any cognitive model of human reasoning.
New information can have a major effect on the reasoning and hence any
computational cognitive model needs to be properly immersed into the external environment,
adapting to new and changing contextual information. Thus in
Cognitive Argumentation the form of arguments and the relation between them
will be context-dependent allowing for the ensuing reasoning via argumentation
to be context-sensitive, e.g.\ influencing which arguments to
consider and the intensity of the process of reasoning.
%

The methodological aim of Cognitive Argumentation is to study different cases of human reasoning
in order to incrementally inform and extend the framework  into an increasingly more general
cognitive model for human reasoning.
The approach of Cognitive Argumentation has already been applied and tested on one such
example concerning how humans reason with Aristotelian Syllogisms~\cite{ki:2019}. This is
an important first case as in this humans are asked to reason as close as possible
to a formal setting and a cognitive model would need to match together the formal and
non-formal human reasoning that is observed to occur. Cognitive Argumentation
performs well in doing this as attested in the recent first Syllogism Challenge 2017.\footnote{\url{https://www.cc.uni-freiburg.de/modelingchallenge/challenge-2017}}

\subsection{Study and Structure of Paper}

In this paper, we will consider a second example of human reasoning, in a very different
setting from that of syllogistic reasoning, namely that of informal or common sense reasoning
with everyday conditional information.
The aim is to formulate within Cognitive Argumentation human conditional reasoning
and test this by examining how it can capture the experimental results of the
suppression task~\cite{byrne:89,dieussaert:2000}, in a coherent and complete
way. The larger aim is to use this study to
probe more deeply the framework of Cognitive Argumentation
in order to understand more generally how to build and apply this
framework.

A large amount of literature exists that investigates the experimental results of the suppression task,
as well as some formal approaches that suggest how to model the task~(e.g.~\cite{stenning:vanlambalgen:2008,cogsci:2012}).
These works mostly concentrate on understanding the suppression effect between the different groups.
Yet, the results contain more information.
In particular, we can identify different kinds of \textit{majorities}, ranging from close to average to almost all participants. 
The experiment thus contains the additional challenge to provide an
explanation for the (significant) majorities and the variation
of conclusions drawn amongst them.
We will see that this is indeed the case in Cognitive Argumentation,
accounting not only for the effect of suppression but also
for these variations among the
population participating in the same groups.
%

The formal framework of Cognitive Argumentation will be presented in Section~\ref{sect:argumentation}.
This will be defined as an instance of preference-based
argumentation~\cite{KMD94,DeLP,SartorPrakkenS97,ModgilPrakken2013,AmgoudDM08,KM03},
suitably adapted for the task of capturing human logical reasoning, and
whose acceptability semantics has the degree of flexibility needed for
the informal nature of human reasoning.
After a brief introduction of the suppression task in
Section~\ref{sect:informal}, we will analyze a set of relevant cognitive principles with particular attention
to the cognitive links between conditionals and argument schemes (Section~\ref{sect:cogprin}).
Section~\ref{sect:bst} presents an analysis of argumentative reasoning of all cases of the suppression task and evaluates 
this in accordance with the observed experimental data, accounting for the suppression effect and the signifigant variation 
of individual responses in the same case.
Section~\ref{sect:cognica} introduces \COGNICA, a web based system for automating the process of human conditional reasoning through Cognitive Argumentation.
The paper ends with a general discussion of human reasoning via argumentation (Section~\ref{sect:discussion}),
summarizing the essential elements of this and the main challenges that lay ahead.

%% file: informal.tex
The \textit{suppression task}~\cite{byrne:89}, is a well-known psychological study on human reasoning.
The experimental setting was as follows: Three groups of participants were asked to derive conclusions given variations of a set of premises. Group I was given the following two premises:\footnote{The participants received the natural language sentences but not the abbreviated notation on the right hand side.}
\begin{align}
\textit{If she has an essay to finish, then she will study late in the library.} \label{e:2} \tag{$\smpl$}\\
\textit{She has an essay to finish.} \label{e:11} \tag{$e$}
\end{align}
The participants were asked what \textbf{necessarily} had to follow assuming that the above two premises were true. They could choose between the following three answer possibilities:\footnote{Here and in the sequel, we will denote with an overbar the negation or complement of an atomic statement, e.g.\ $\ness$ and $\nlib$ denote the negation of $e$ and the negation of $\lib$, respectively.}
\begin{align}
\textit{She will study late in the library.} \label{l:1} \tag{$\lib$}\\
\textit{She will not study late in the library.} \label{l:2} \tag{$\nlib$}\\
\textit{She may or may not study late in the library.} \label{l:3} \tag{$\lib$ or $\nlib$}
\end{align}
In this first group, 96\% of the participants concluded that \textit{She will study late in the library}.

In addition to the above two premises for Group~I, Group~II was given the following premise:
\begin{align}
\textit{If she has a textbook to read, then she will study late in the library.} \label{e:3} \tag{$\altn$}
\end{align}
Still, 96\% of the participants concluded that \textit{She will study late in the library}.
Finally, Group~III received, together with the two premises of Group~I, additionally the following premise:
\begin{align}
\textit{If the library stays open, then she will study late in the library.} \label{e:4} \tag{$\add$}
\end{align}
In this group only 38\% concluded that \textit{She will study late in the library}: The conclusion drawn in the previous groups was \textit{suppressed} in Group~III. 

The results of this experiment show that previously drawn conclusions seem to be suppressed given (appropriate) additional information, i.e.\ participants seemed to reason non-monotonically \textbf{in a context-sensitive way}. A natural explanation why participants in Group~III did not conclude necessarily that \textit{She will study late in the library}, is because they
were not sure whether \textit{The library stays open}, which is a necessary requirement for her to study late in the library. 
In the first two groups the majority of the participants did not
have this doubt, as they had not been made \textit{aware} of the possibility that the library may not be open.

 \begin{figure}\centering
\begin{tikzpicture}[scale=0.1,-latex ,auto ,node distance =1 cm and 2cm ,on grid , semithick ,state/.style ={ circle, minimum width =0.5 cm}]

\node[state,align=center,rectangle,fill=\mycolorG] (A1) {\small Fact: $e$, $e \imply \ell$};

\node[state,align=center,rectangle,fill=\mycolorR] (A5) [above =+1cm of A1] {Hyp: \small $\overline{\ell}$};

\node[state,align=center,rectangle,fill=\mycolorG] (A6) [above =+1cm of A5] {Hyp: \small $\ell$};

\draw[->] (A1) edge node[left] {}  (A5);
\draw[->] (A5) edge[bend right] node[left] {} (A6);
\draw[->] (A6) edge[bend right] node[left] {} (A5);

\node[state,align=center,rectangle,fill=\mycolorG] (A21) [right =+3.5cm of A1]{\small Fact: $e$, $e \imply \ell$};
\node[state,align=center,rectangle,fill=\mycolorG] (A23) [right =+3cm of A21] {\small Hyp: $t$, $t \imply \ell$};

\node[state,align=center,rectangle,fill=\mycolorR] (A25) [above right=+1cm and 1.3cm of A21] {Hyp: \small$\overline{\ell}$};
\node[state,align=center,rectangle,fill=\mycolorG] (A26) [above =+1cm of A25] {Hyp: \small$\ell$};

\draw[->] (A21) edge node[left] {} (A25);
\draw[->] (A23) edge node[left] {} (A25);

\draw[->] (A25) edge[bend right] node[left] {} (A26);
\draw[->] (A26) edge[bend right] node[left] {} (A25);

\node[state,align=center,rectangle,fill=\mycolorG] (A31)[right =+6.5cm of A21] {\small Fact: $e$, $e \imply \ell$};

\node[state,align=center,rectangle,fill=\mycolorG] (A33) [below =+1cm of A31] {\small Hyp: $\overline{o}$, $\overline{o} \imply \overline{\ell}$};

\node[state,align=center,rectangle,fill=\mycolorG] (A35) [above =+1cm of A31] {\small Hyp: $\overline{\ell}$};
\draw[->] (A31) edge[bend left] node[right] {} (A33);
\draw[->] (A33) edge[bend left] node[right] {} (A31);

\node[state,align=center,rectangle,fill=\mycolorG] (A36) [above =+1cm of A35] {\small Hyp: $\ell$};


\draw[->] (A31) edge node[left] {} (A35);

\draw[->] (A35) edge[bend right] node[left] {} (A36);
\draw[->] (A36) edge[bend right] node[left] {} (A35);

\end{tikzpicture}

\caption{\label{fig:part1:e:intro} Summary of the case where \textit{She has an essay to finish} for group~I (left), group~II (middle) and group~III (right).
The acceptable arguments are highlighted in gray.}
\end{figure}
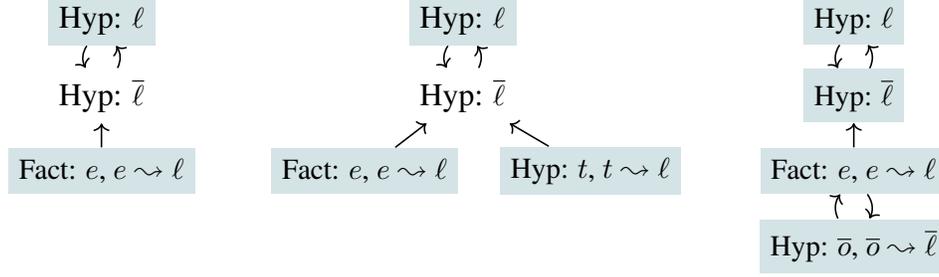 
In this paper, we will show how this experiment and its observed data can be naturally understood by formalizing human reasoning in terms of building supporting arguments for conclusions and defending such arguments against their 
counterarguments.

To reason in terms of argumentation, we can construct an argument based on $e$ and $\smpl$, which supports the conclusion $\lib$. 
In Groups I and II, the only argument that we can construct for $\nlib$ consists of hypothesizing $\nlib$ itself. This forms a counterargument to the above argument for $\lib$. But according to a relative preference or strength of the explicitly stated premises, the argument of $e$ and $\smpl$, can defend against $\nlib$, but not vice-versa
as $\nlib$ is a weaker argument. The left graph in Figure~\ref{fig:part1:e:intro} representing Group~I, shows at the bottom this winning argument. To distinguish explicitly stated premises not those not stated explicitly we will call the latter 
\textit{hypothetical} premises denoting them with a prefix of `Hyp'.
At the top of the figure we see another argument supporting $\lib$, namely the 
 hypothesis of $\lib$. The figure shows how this is attacked by hypothesis argument for $\nlib$ and that how $\lib$ can defend back against this as there is no preference (or strength) between these two hypotheses. 
 Given that the hypothesis argument for $\nlib$ cannot defend against its counterargument, $e$ and $\smpl$, we have a good quality, or \textit{acceptable} as it is normally called in AI, argument for $\lib$ but not for $\nlib$.
 
The middle part of Figure~\ref{fig:part1:e:intro}  shows the case for Group II where we can construct another argument for $\lib$ based on the hypothesis of $\tbook$ and $\altn$.
In both groups, $\lib$, is the only statement supported by acceptable arguments. This corresponds to the majority's conclusion, that \textit{She will study late in the library} holds, i.e.\ that this is a \textbf{definite conclusion}.

For Group III, the case is different. The participants might have become aware of the common sense knowledge that \textit{If the library does not stay open, then she will not (be able to) study late in the library} ($\negadd$). This together with the possibility, or the hypothesis, that \textit{The library does not stay open}~($\overline{o}$), gives an argument that supports the conclusion that $\nlib$. Furthermore, it seems that this argument is at least as preferred as the argument supporting $\lib$ based on $e$ and $\smpl$. Hence now we also have an argument for $\nlib$ that is acceptable and those human participants that reason with this argument are prevented from deriving that \textit{She will study late in the library} as a \textbf{definite conclusion.} For them, it is only a  \textbf{plausible conclusion.}
Figure~\ref{fig:part1:e:intro} illustrates this in the right-most part where although the same argument, $e$ together with $\smpl$, continues to defeat the hypothetical argument for $\nlib$ it does not defeat the argument built from $\overline{o}$ and $\negadd$, as seen at the bottom of the figure. Both
these arguments can defend against each other and hence they are both acceptable.



In total, \cite{byrne:89} reported the experimental results of twelve cases of the suppression task.
For each of the three groups, four different cases of reasoning were considered by combining their general knowledge with one of the following factual information:
\begin{align}
\textit{She has an essay to finish.}  \tag{$e$}\\
\textit{She will study late in the library.} \label{e:51} \tag{$\lib$} \\
\textit{She does not have an essay to finish.} \label{e:12} \tag{$\ness$}\\
\textit{She will not study late in the library.} \label{e:52} \tag{$\nlib$} 
\end{align}
Table~\ref{tab:summary-suppression} shows for each group (column 1), the conditional information they received (column 2) together with the factual information for each of the four cases (column 3 to 6). 
In each row we can see the percentage of responses by the participants in the
group corresponding to the row of the table. Those in gray are the responses 
demonstrating the suppression effect.
The majority's responses in Group II diverges in two cases from the majority's responses in Group I and III: When participants received
the information, that \textit{She does not have an essay to finish} ($\ness$), only 4\% concluded that \textit{She will not study late in the library} ($\nlib$), and when they received the information that 
\textit{She will study late in the library}, only 13\% concluded that \textit{She has an essay to finish}.
Contrary to these cases, the suppression effect for Group III took place when participants received the information
that \textit{She has an essay to finish} (only 38\% concluded that \textit{She will study late in the library}), or
when they were given the fact that \textit{She will not study late in the library} (only 33\% concluded that \textit{She does not have an essay to finish}.

\begin{table}
\begin{center}
\begin{tabular}{@{\hspace{0cm}}l@{\hspace{0.5cm}}l@{\hspace{0.5cm}}c@{\hspace{0.3cm}}c@{\hspace{0.3cm}}c@{\hspace{0.3cm}}c}
\toprule
Group  & Conditional(s) & Essay ($e$) & No essay ($\ness$) & Library ($\lib$) & No library ($\nlib$)  \\\midrule\midrule 
I & $\smpl$ &  library {\color{black}(96\%)} & no library {\color{black}(46\%)} & essay (71\%) & no essay (92\%) \\  \midrule 
II & $\smpl$, $\altn$ & {library {\color{black}(96\%)}} & {{\color{gray}no library (4\%)}}
& {{\color{gray}essay (13\%)}} &  {no essay (96\%)}
\\  \midrule 
III & $\smpl$, $\add$ & {{\color{gray}library (38\%)}} & 
{no library {\color{black}(63\%)}} & {essay (54\%)} & {{\color{gray}no essay (33\%)}} \\\bottomrule
\end{tabular}
\caption{Summary of the twelve cases and the corresponding suppression effects denoted in gray.\label{tab:summary-suppression}}
\end{center}

\end{table}

We will use this experiment to motivate and test our model of Cognitive Argumentation for human reasoning by examining how it can uniformly capture the experimental results in all twelve cases
accounting for the suppression effect as well as the variation of responses within each group.

%% file: cognitiveprinciples-short.tex
Humans make various (implicit) assumptions while reasoning, many of which are not necessarily valid under (formal) classical logic. We will specify such (typically) extra-logical properties and formalize them as \textit{cognitive principles} to help us develop  a framework of argumentation that is in tune with human reasoning.

\subsection{Maxim of Quality}\label{sect:maxqual} 

According to Grice's~\citeyear{grice:1975} conversational implicature, humans communicate according to a cooperation principle.
The maxim of quality states that humans try to be truthful and thus information that we get in conversation is assumed to be true and trusted. In the context of the suppression task, for example,
this principle implies that what the experimenter states, 
e.g.\ \textit{She has an essay to finish} ($\essay$), is believed to be true by the participants: it is trusted as strong information that does not need to be questioned or is questioned only in an extreme case.

Accordingly, we will establish a (strong) factual argument scheme to encompass this principle. 

\subsection{Maxim of Relevance}\label{sect:maxrel}  

People consider different scenarios depending on whether they have been made \textit{aware} of alternative options while reasoning~\cite{sperber:wilson:1995,byrne:2005}. This awareness
may not be through some direct and explicit mention of the alternative. Nevertheless, considering Grice's~\citeyear{grice:1975} maxim of relevance, it seems natural to consider (and account) for the possibility of these alternatives, as the participants might believe that, otherwise this information would not have been mentioned, e.g.\ in a dialogue.
We can capture this cognitive principle of considering different awareness (or relevance) driven possibilities through a (weak) hypothesis argument scheme.


Hence for information that we are made aware of and not given explicitly as factual information, people can still construct various context-dependent hypotheses supporting statements concerning this information. As there is no direct evidence that these hypotheses hold, they are only plausible, the hypothesis argument scheme is weaker than other argument schemes based on explicitly given information.

\subsection{Conditional Reasoning} \label{sect:conditionals}


Byrne~\citeyear{byrne:2005} distinguishes between different types of conditionals and conditions, assumed to be perceived by humans in situations like those of the 
selection task~\cite{wason:68,griggs:cox:1982} and the suppression task~\cite{dietz:hoelldobler:rocha:2017}. We will extend this distinction and propose canonical associations related to different types of conditions. In particular, we will introduce and distinguish between prediction and explanatory associations and related argument schemes.

Consider the following conditional:

\begin{align}
\textit{If I need milk, then I will buy milk.} \label{needmilk} \tag{$\needmilk$} 
\end{align}
The condition \textit{I need milk} can be understood as \textbf{sufficient}, in the sense that if the condition holds, 
then this forms a support for the consequent, \textit{I will buy milk}, to hold as well (modus ponens). On the other hand,
the negation of the condition, \textit{I don't need milk}, seems to be a plausible support for the negation
of the consequent, \textit{I will not buy milk} (denying the antecedent). Thus the condition can also be understood as
\textbf{necessary} for the consequence to hold.

Consider now, additionally to~(\ref{needmilk}), the following conditional:
\begin{align}
\textit{If my mother asks me to get her milk, then I will buy milk.} \label{asksmilk} \tag{$\asksmilk$} 
\end{align}
Both conditions in~(\ref{needmilk}) and~(\ref{asksmilk}), which are independent of each other, are separately sufficient in order 
for the consequence to hold. 
However, the negation of either of these conditions alone is not enough
 to conclude the negation of the consequence, \textit{I will not buy milk}. Only the negation of both conditions together, gives sufficient support
  to conclude the negation of the consequence. 
  Therefore, individually the conditions in (\ref{needmilk}) and~(\ref{asksmilk}) are not necessary conditions. Now that there is a second way to bring about the consequent, the condition \textit{I need milk} has lost its necessary property. 

Let us now assume that, in addition to~(\ref{needmilk}) and~(\ref{asksmilk}), we are also given the following conditional:
\begin{align}
\textit{If I have enough money, then I will buy milk.} \label{moneymilk} \tag{$\moneymilk$}
\end{align}
 By this conditional~(\ref{moneymilk}) we are made aware of the possibility of a situation where, even in the case where, \textit{I need milk}
 or \textit{my mother asks me to get her milk}, I might not buy milk, because possibly \textit{I don't have enough money}.
 Having enough money is a necessary condition for the consequent: without it the consequent cannot hold, i.e.\ \textit{I cannot buy milk}, no matter what other (sufficient) conditions might hold at the time. Also in comparison with the 
 the above cases we might consider this a \textbf{strong necessary} condition in the 
 sense that it is more or very unlikely for this to loose its necessary property.
 On the other hand, the condition of~(\ref{moneymilk}) cannot be considered as a sufficient condition: even if \textit{I have enough money}, I might not buy milk.

The distinction between the two different types of conditions, sufficient and necessary, shows up when we consider \textbf{explanations} of the consequent and its negation. 
Assume that we are given the information that
\begin{align}
 \textit{I did not buy milk.} \label{dontbuy} \tag{$\didntbuy$} 
\end{align}
It is reasonable that, given~(\ref{needmilk}) and~(\ref{asksmilk}) (without (\ref{moneymilk})), to conclude the negation of the condition of both conditionals, namely that \textit{I did not need milk} \underline{and} \textit{my mother did not ask me to get her milk} (modus tollens).
Adding the conditional~(\ref{moneymilk}) in the context of reasoning would not extend this conjunction but would result in a disjunctive addition of the negation of the new (necessary) condition:
\begin{center}
Either \hfill (\textit{I do not need milk} \underline{and}  \textit{my mother does not ask me to get her milk})\\ \underline{or} \hfill  \textit{I do not have enough money}.
\end{center}

Hence the observation of the negation of the consequent can be \textbf{explained} by the negation of a necessary condition (e.g. 
\textit{I do not have enough money}) or by assuming that there is ``no reason'' for the consequent to hold, resulting in a more complex explanation, namely that none of the sufficient conditions can hold.
%

In contrast, if we are given the positive information that a consequent holds, e.g. \textit{I buy milk}, then this can be simply \textbf{explained} by any one of the sufficient conditions for the consequent, e.g. either by 
\textit{I need milk} or by \textit{my mother asks me to get her milk} (affirming the consequent).
It is important to note that typically we will not consider that two such sufficient conditions, together, form an explanation. In fact, we typically consider that different explanations are incompatible with each other, except perhaps in very exceptional cases where many different reasons can hold together. Hence we will only accept \underline{one}, either \textit{I need milk} or \textit{my mother asks me to get her milk} to explain the consequent \textit{I buy milk} but not both together. Similarly, when we are explaining the negation of the consequent, e.g. \textit{I did not buy milk}, we will only accept \underline{one} of the explanations, either \textit{I do not have enough money} or there is ``no reason'', i.e. \textit{I did not need milk} \underline{and}  \textit{my mother did not ask me to get her milk}. 

Hence different explanations are in general considered to be in tension with each other. They are competing or contrasting alternatives as implied for example by the maxim of ``Inference to the best explanation'' (see e.g. \cite{LiptonBook,RubenBook}). The process of explanation is not merely to find why something holds but also why this is indeed the reason for holding and not for some other reason. In \cite{kelley:1973,Sloman94} a 
cognitive principle of \textbf{explanatory discounting} is identified which assumes that alternative explanations are in conflict with each other so that support for one explanation results in diminishing support, thus countersupport, against alternative explanations. 

Human explanatory reasoning also follows  a principle of \textbf{simplicity} by choosing simple explanations depending on the context at hand. Hence even when a ``logically complete'' explanation would contain 
a conjunction, such as the explanation of \textit{I did not need milk} \underline{and}  \textit{my mother did not ask me to get her milk}, we would consider these conjuncts as separate explanations drawn from the observation, depending on the context of reasoning. In one context, when we are buying milk for ourselves, we would explain not buying milk by \textit{I did not need milk} and conclude this, without necessarily also considering the explanation \textit{my mother did not ask me to get her milk}.

Finally, we note that depending on the nature of the condition, sufficient or necessary, we can draw conclusions in a \textbf{secondary predictive mode} from factual information about the consequent.  Observing the negation of the consequent can lead us to predictively conclude the negation of any of its sufficient conditions.
This can be seen as a ``modus tollens'' conclusion via Reductio ad Absurdum based on the fact that when a sufficient condition holds its consequent also has to hold.
Furthermore, in some cases these conclusions can be the same as conclusions drawn in a \textbf{secondary explanatory mode} by considering that the negation of a sufficient condition is an explanation of the observed failure of the consequent to hold in the particular context in which we are reasoning.

On the other hand, observing that the consequent holds can lead us to predictively conclude that a necessary condition holds, e.g. observing \textit{I buy milk}, we can conclude that \textit{I have enough money}. This follows from the way a necessary condition is understood, i.e.\ \textit{I have enough money} must necessarily hold for the consequent to hold (affirming the consequent). 
%

As discussed previously, cognitively,  an explanation is required to discriminate between different possible alternatives. Howevever, a necessary condition cannot be considered as a possible explanation for the consequent because it always holds and hence it does not offer any \textit{discriminatory information}. 
Prediction of several necessary conditions, as opposed to different explanations, do not compete with each other~\cite{FernbachSloman2010}
and hence they can hold together when we are given that the consequent holds.

Summarizing, we note that the interpretation of conditions within conditionals as sufficient or necessary and the possible conclusions drawn either as predictions or explanations depends on the context in which these are considered.
 This \textbf{context-sensitive} nature of interpretation and reasoning can vary among the population, depending on the background knowledge of conditionals that each individual has or that is made aware of in a scenario of discourse.


\subsection{Canonical Associations of Condition and Consequence}\label{sect:conditionals}

We will now establish canonical associations of
different types of conditions with respect to predictions and explanations, which in turn will 
correspond to argument schemes that will form the basis for the argumentative reasoning. 
Consider a {\condition} and a {\consequence} coming from some conditional:
\text{``if {\condition} then {\consequence}''.} 
\begin{table}
 \begin{tabular}{@{\hspace{0cm}}lllcccc} \toprule
  Type & Applicable Principle & Abbrev  & \multicolumn{4}{c}{Given Fact} \\
                    &                       & & \cond & $\ncond$  & \consq & $\nconsq$ \\\midrule
  $\suf$   & \cogPrinPsuff & (\ref{psuff})& \consq & - & - & - \\
  \cmidrule{2-7} 
  &                 \cogPrinEsuff &  (\ref{esuff}) & - & - & \cond & - \\ 
  & \cogPrinsecPsuffabbr & (\ref{secpsuff})       & - & - & - & $\ncond$ \\
  & \cogPrinsecEsuffabbr &  (\ref{secesuff})       & - & - & - & $\ncond$ \\\midrule 
  $\nec$  &  \cogPrinPnecc & (\ref{pnecc})        & - & $\nconsq$ & - & - \\
  \cmidrule{2-7} 
  & \cogPrinsecPneccabbr & (\ref{secpnecc})        & - & - & \cond & - \\
  & \cogPrinEnecc& (\ref{enecc})        & - & - & - & $\ncond$ \\  \midrule
  & \cogPrincExo & (\ref{exo}) & - & - &  $\exo(\consq)$   &  $\exo(\nconsq)$  \\   \bottomrule 
 \end{tabular}
\caption{Predictions and explanations from factual information depending on the type of condition.
\label{tab:expl}}
\end{table}

We establish the following
rule associations\footnote{Associations are written with $\imply$ instead of $\rightarrow$ to emphasize their defeasible nature.} between a condition and a
consequent:
\begin{enumerate}
\item\label{prediction} \textbf{Predictions:} The canonical predictive association for a sufficient condition:
\begin{align}
\cond \imply \consq \tag{\psuff} \label{psuff}
\end{align}
The canonical predictive association for a necessary condition:
\begin{align}
\overline{\cond} \imply \overline{\consq} \tag{\pnecc} \label{pnecc} 
\end{align}
\item\label{explanation} \textbf{Explanations:} The canonical explanatory association
for a necessary condition:
\begin{align}
\overline{\consq} \imply \overline{\cond} \tag{\enecc} \label{enecc}
\end{align}
The canonical explanatory association for a sufficient condition:
\begin{align}
\consq \imply \cond \tag{\esuff} \label{esuff} 
\end{align}
Note that these explanatory associations are the reverse of the predictive ones. 
\item\label{secondary} \textbf{Secondary Associations:}
\begin{enumerate}
 \item\label{secondary:prediction} \textbf{Secondary Predictions:}
 The secondary association (which corresponds to the \textit{contrapositive} of the prediction association)
for a sufficient condition, is:
\begin{align}
\overline{\consq} \imply \overline{\cond} \tag{\secpsuff} \label{secpsuff} 
\end{align}
The secondary association for a necessary condition is:
\begin{align}
\consq \imply \cond \tag{\secpnecc} \label{secpnecc} 
\end{align}
 \item\label{secondary:explanation} \textbf{Secondary Explanations:}
The {secondary explanatory} association for a sufficient condition is:
\begin{align}
\overline{\consq} \imply \overline{\cond} \tag{\secesuff} \label{secesuff}
\end{align} 
 \item\label{exognous:explanation} \textbf{Exogenous Explanations:}
Psychological experiments (e.g.~\cite{exo:reasons:2010}) show that humans are sometimes likely to come up with alternative causes that are \textit{not appearing} within the given context. These exogenous explanations can be captured via associations which link the consequent (positive or negative) with an exogenous explanation (represented by $\exo(\consq)$ and $\exo(\overline{\consq})$ respectively).
\begin{align}
\consq \imply \exo(\consq) \quad\quad\quad\quad \mbox{and} \quad\quad\quad\quad
\overline{\consq} \imply \exo(\overline{\consq}) \tag{\suffexo} \label{exo}
\end{align}
 \end{enumerate}
\item\label{strength} \textbf{Strength of Associations:}
Predictive associations from necessary conditions (\ref{pnecc}) are stronger than conflicting associations from sufficient conditions (\ref{psuff}).
This reflects the strength of a pragmatic \textit{disabling condition} over a
motivational \textit{enabling condition} for the same consequent.
\item\label{incompatibility} \textbf{Incompatibility:}
Explanatory associations are typically incompatibly exclusive. For example, if there is more than one explanatory sufficient condition for the consequence  then they are incompatible with each other and of equal strength. Table~\ref{tab:incompatibility} provides a summary of the (in)compatibility of explanations. Note that exogenous explanations are by their nature in conflict with other explanations: people introduce them only when they are in doubt about other explanations.
\end{enumerate}


Following the discussion above, Table~\ref{tab:expl} gives a summary of the condition types together with their canonical predictions and explanations
with respect to the given facts.
It is important to note that these canonical predictions and explanations are not meant to necessarily represent definite conclusions but rather that they are plausible conclusions that are cognitively admissible in human reasoning. 

\begin{table}\centering
\begin{tabular}{@{\hspace{0cm}}l@{\hspace{0.2cm}}lG}
Observation & (In)compatibility of explanations & CL\\\midrule
 $\consq$ & Explanations from sufficient conditions mutually incompatible  & - \\ 
$\overline{\consq}$  &  Explanations from  sufficient conditions mutually compatible & Contraposition \\
$\overline{\consq}$   &  Explanations from  necessary conditions mutually incompatible & Contraposition\\
&  Also incompatible with explanations from sufficient conditions  \\
 $\consq$ & Exogenous explanation incompatible with any other explanation  & - \\ 
 $\overline{\consq}$ & Exogenous explanation incompatible with any other explanation & - \\ 
\end{tabular}
\caption{
Overview of the (in)compatibility of explanations given the observation.
\label{tab:incompatibility}}
\end{table}

%

\begin{table}\begin{center}
              
\begin{tabular}{@{\hspace{0cm}}G@{\hspace{0cm}}l@{\hspace{0.7cm}}l@{\hspace{0.7cm}}G@{\hspace{0cm}}l@{\hspace{0.7cm}}G@{\hspace{0cm}}l@{\hspace{0cm}}G}
\toprule Cognitive Principle &  & \centering Group I & Prediction & {\centering  Group II} & Prediction &\centering  Group III & Prediction \\\midrule
\cogPrinPsuff & (\ref{psuff}) & $\smpl$ &   {$\ell$}  & $\smpl$, $\altn$  &  {$\ell$}         & $\smpl$    & - \\  
      \cogPrinPnecc     &           (\ref{pnecc}) & ($\negsmpl$) &  &  & & $\negadd$, ($\negsmpl$) &  \\\midrule

\cogPrinsecPnecc & (\ref{secpnecc}) & ($\bismpl$) & {$e$}  &    &           & ($\bismpl$), {$\biadd$}    &  \\  
 \cogPrinsecPsuff       &              (\ref{secpsuff})  & $\negbismpl$ &    & $\negbismpl$, {$\negbialt$} & - & $\negbismpl$ & {$\nlib$} \\ \midrule
\cogPrinEsuff        &     (\ref{esuff}) & $\bismpl$ &  {$\ness$} & $\bismpl$, $\bialtn$ & {$e \vee \tbook$} & $\bismpl$ &  \\
\cogPrinEnecc & (\ref{enecc}) & ($\negbismpl$) & {$\nlib$} &    &          & ($\negbismpl$), $\negbiadd$    &  \\ \midrule
 \cogPrinsecEsuff       &              (\ref{secesuff})  & $\negbismpl$ &    & $\negbismpl$, {$\negbialt$} & - & $\negbismpl$ & {$\nlib$} \\ \midrule 
 &  (\suffexo) &  
$\lib \imply {\tiny\exo(\lib)}$,
$\nlib \imply {\tiny\exo(\nlib)}$ & &
$\lib \imply {\tiny\exo(\lib)}$,
$\nlib \imply {\tiny\exo(\nlib)}$ & & 
$\lib \imply {\tiny\exo(\lib)}$,
$\nlib \imply {\tiny\exo(\nlib)}$
 \\\bottomrule
\end{tabular}
             \end{center}

\caption{Classification of the conditions of the conditionals of the suppression task.\label{tab:can:bst:I}}
\end{table}
Table~\ref{tab:can:bst:I} shows, as an example, the canonical associations that apply for the suppression task. 
For the cases of Group I and III the condition \textit{She has an essay to finish} can be interpreted both as sufficient and necessary. For some part of the population this may only be a sufficient condition in which case the associations shown in parentheses in the columns for Groups I and III will not apply. 
In Group II, \textit{She has an essay to finish} is no longer considered as necessary due to the presence of a second sufficient condition of \textit{She has a textbook to read}. They are both considered in the whole population of Group~II only as sufficient conditions.

%% file: argumentation.tex
We will now define the formal framework of Cognitive Argumentation for human reasoning. 
This framework will encompass the conditional associations and other cognitive principles as argument schemes together with their strength relation. 
It will be based on the standard framework of abstract argumentation in AI, ~\cite{dung95}, as realized in preference based structured argumentation (see e.g.~\cite{KMD94,SartorPrakkenS97,KM03,ModgilPrakken2013}).
Before presenting its technical definition we will first summarize, informally, its essential elements and how logical reasoning is realized through argumentation. 

\subsection{Reasoning via Argumentation}\label{InformalArg}

In argumentation the essential and general structure of knowledge is an \textit{argument scheme}: a structure that simply associates
or links two pieces of information, the premises with the claim or position of the scheme.
Arguments are built from (argument) schemes, and support the corresponding particular instances of the positions of these schemes.
Then reasoning in argumentation is a process of analysis of alternatives,
e.g.\ a conclusion and its negation, by a consideration of different arguments
for and against the various competing alternatives. In comparison with
other classical logical approaches, reasoning via argumentation is an explicit process of
examining the alternatives either at the level of the final conclusion we are
interested in or at the level of other information, e.g.\ premises
of arguments, that support the alternative possible conclusions.

More concretely, in a framework of argumentation-based reasoning the \emph{essential structural elements} are, the argument schemes, a notion of relative conflict and a notion of relative
strength between argument schemes and thus between arguments formed from them. Reasoning is then
a process of \emph{dialectic argumentation} where starting from an argument supporting
a position of interest we consider arguments that, under the comparative
conflict relation, compete, e.g.\ arguments supporting incompatible positions,
and examine how we can defend against such counterarguments through arguments
which are stronger or at least not weaker than the counterarguments. 

Arguments that can defend themselves against their counterarguments are called \emph{acceptable} and the positions that they support are \emph{plausible} conclusions. When in addition no such acceptable argument can be constructed for the complement of a position then this forms a
\emph{definite} conclusion. Definite conclusions correspond to logical conclusions in formal systems of logic: they are certain and undisputed. On the other hand, plausible conclusions are useful in informal human reasoning indicating the possibility of holding. 
We can therefore notice that in its general form, reasoning via argumentation is close to model construction, similar as in mental model theory~\cite{johnsonlaird:1980,johnsonlaird:2004}.

Any framework for human reasoning when
placed into practise needs to be influenced by other, possibly extra-logical, factors that
play a significant role in the reasoning process by giving it a \textit{cognitive form}. 
In computational terms we can
think of this as cognitive-based heuristics that humans have learned to use to make their reasoning effective.
Examples of such extra-logical factors that govern the shape of human
reasoning when this is understood via argumentation are: awareness of the
relevant argument schemes in the current context of reasoning, recognition
and selection of relatively strong arguments and the bounded application of the
dialectic process by concentrating only on some of the possible counterarguments.
%
%
The challenge is to apply the process of dialectic argumentation in a dynamic and context-sensitive way that focuses on the parts of the knowledge pertinent to the reasoning task at hand.

\subsection{Argumentation Logic Framework} \label{argschemes} \label{sub:alf}

Within the formal argumentation logic framework, the atomic statements in natural language can be represented as propositional variables.\footnote{For simplicity of presentation we will only consider propositional argumentation frameworks. }
%
Given a propositional logical language, $\CalL$, the \defname{set of propositional variables} 
in $\CalL$ is denoted by {$\CalP_\CalL$}. 
For the ``milk example'' of Section~\ref{sect:conditionals}, 
$\CalP_\varmilk$, consists of variables representing \textit{I need milk}, \textit{I will buy milk},
 \textit{My mother asks me to get her milk} and \textit{I have enough money}, as follows:
\[\begin{array}{lll}
   \CalP_\varmilk & =& \{
   \varneedm,\ \varasksm,\ \varbuym,\ \varmoneym \}.
   \end{array}
\]

The negation of $\CalP_\CalL$, is denoted by $\neg \CalP_\CalL = \{ \overline{x} \mid x \in \CalP \}$,
where $\overline{x} = \neg x$. In general, $\overline{L} = \neg A$ if $L = A$
and $\overline{L} = A$ if $L = \neg A$.
Accordingly, the negation of $\CalP_\varmilk$ is given by:
\[\begin{array}{lll}
   \neg \CalP_\varmilk & =&  \{
   \nvarneedm,\ \nvarasksm,\ \nvarbuym,\ \nvarmoneym \}.
  \end{array}
\]
We will be interested in reasoning within a \textbf{cognitive state $\CalS = (\FPrem, \HPrem)$}  where, $\FPrem$, is a \textit{set of facts} and $\HPrem$ is an \textit{awareness set}. Both these elements of a cognitive state are linked to the environment of the reasoner, the first consisting of explicit factual information that the environment provides while the second consists of the concepts that the reasoner is made aware of by the environment. We have $\FPrem \subseteq (\CalP_\CalL \cup \neg \CalP_\CalL)$ and $\HPrem \subseteq \CalP_\CalL$. Note that the awareness of concepts in $\HPrem$ does not necessarily mean that we are aware whether they hold or not but simply that they and knowledge about them might be relevant to the reasoning at hand. Also any concept for which we have a fact in $\FPrem$ belongs to $\HPrem$,
i.e.\ if $A \in \FPrem$ or $\overline{A} \in \FPrem$, then $A \in \HPrem$.

Given a propositional language, $\CalL$, an \textbf{argumentation logic framework} is a triple $\ALF = \langle \ASset, \Cf, \St \rangle$ where
$\ASset$ is a set of argument schemes, $\Cf$ is a conflict relation on $\ASset$, typically induced by the notion of conflict in the language~$\CalL$, and~$\St$ is a binary strength relation on $\ASset$.
An \textbf{(argument) scheme}, $\AS \in  \ASset$, is a tuple of the form $\AS =  (\Pre, \Pos)$ where \defname{precondition \Pre} and \defname{position \Pos} are (sets of) statements in the language~$\CalL$.
For instance, an argument scheme will link subsets of propositional variables, i.e.\ $\Pre, \Pos \subseteq (\CalP_\CalL \cup \neg \CalP_\CalL)$.

Argument schemes were introduced as stereotypical reasoning patterns that are typically non-deductive and
non-monotonic~\cite{pollock:1995,walton:1996}.
In general, they allow us to link the information in $\Pre$ with that of $\Pos$. 
Usually argument schemes are parametric
so that a scheme can be applied for different values of its parameters to give or to construct arguments. 
We normally say that $\Pre$ are the premises on which the position $\Pos$ is supported by an argument constructed through the scheme $\AS =  (\Pre, \Pos)$. 
We will use argument schemes to capture the canonical associations motivated by the cognitive principles in
Section~\ref{sect:cogprin}.  

Recall the principle of \textit{Maxim of Quality} in Section~\ref{sect:maxqual}, under which what is given as premise (e.g. by the experimenter), is taken to hold.
Accordingly, we introduce the following \textit{fact scheme}:
\[
\begin{array}{l@{\hspace{1cm}}ll}
\eFact(L) =(\emptyset, L) \in \ASset &  \mbox{ if } & L \in \FPrem.
\end{array}
\]
Given the requirement that $L$ needs to be in $\FPrem$, the fact scheme is only applicable when indeed $L$ is a fact in the cognitive state.
%
Similarly, for the principle of \textit{Maxim of Relevance} in Section~\ref{sect:maxrel},
where everything we are made aware of, can possibly hold or not, we introduce 
a corresponding \textit{hypothesis scheme} as follows:
\[
\begin{array}{lll@{\hspace{1cm}}ll}
\ehyp(A) = ( \emptyset, A)  \in \ASset & \mbox{ and } & \ehyp(\overline{A}) = ( \emptyset, \overline{A})  \in \ASset &  \mbox{ if } & A \in \HPrem.
\end{array}
\]
Here we require that the concept referred to in $A$ or $\overline{A}$ needs to appear in the \textit{awareness set} in order for the argument scheme to be applicable: once we are aware of a concept we can hypothesize that it holds or that it does not hold.

In Section~\ref{sect:conditionals} we have
presented how humans consider different types of conditions in relation to a consequent of interest and thus different associations between them. 
The canonical prediction and explanation associations
as summarized in Table~\ref{tab:expl}, are straightforwardly represented by 
the following argument schemes:
\[
\begin{array}{rll@{\hspace{3cm}}rll}
\eASpsuff & = & (\cond, \consq)  & \eASesuff & = & (\consq, \cond)   \\ 
 \eASsecpsuff & = & (\nconsq, \ncond) &\eASsecesuff & = & (\nconsq, \ncond)  \\
\eASpnecc & = & (\ncond, \nconsq)  &  \eASenecc & = & (\nconsq, \ncond) 
\\
\eASsecpnecc & = & (\consq, \cond)  &  \suffexo(\consq) & = & (\consq, \exo(\consq))\\
& & &  \suffexo(\nconsq) & = & (\nconsq, \exo(\nconsq)) 
\end{array}
\]
Depending on the given conditionals and their type of condition, the 
respective schemes will be in $\ASset$, and applicable for the construction of arguments.

Consider the milk example from Section~\ref{sect:conditionals}:
For the two conditionals with sufficient condition
\begin{center}
 (\ref{needmilk})  \quad \quad and \quad\quad  (\ref{asksmilk})
\end{center}
the following two prediction schemes apply:
\[
 \begin{array}{l@{\hspace{0.2mm}}l@{\hspace{0.2mm}}l@{\hspace{0.2cm}}l@{\hspace{0.2cm}}l@{\hspace{0.2mm}}l@{\hspace{0.2mm}}l}
  \ASpsuff{\needmilk} & = & (\varneedm, \varbuym) & \mbox{ and } &
   \ASpsuff{\asksmilk} & = & (\varasksm, \varbuym).
 \end{array}
\]
For the conditional with the necessary condition,
\begin{center}
(\ref{moneymilk})
\end{center}
where \textit{I have enough money} is a necessary condition for \textit{I will buy milk},
and thus, if \textit{I don't have enough money}, then \textit{I will not buy milk}, 
 the following prediction scheme applies:
\[
 \begin{array}{lll@{\hspace{1cm}}lll@{\hspace{1cm}}lll}
  \ASpnecc{\nmoneymilk} & = & (\nvarmoneym, \nvarbuym)
 \end{array}
\]
We also have explanation schemes from the explanation associations, e.g.:
\[
 \begin{array}{l@{\hspace{0.2mm}}l@{\hspace{0.2mm}}l@{\hspace{0.2cm}}l@{\hspace{0.2cm}}l@{\hspace{0.2mm}}l@{\hspace{0.2mm}}l}
  \ASesuff{\buyneed} & = & (\varbuym, \varneedm) & 
  \ASenecc{\nbuymoney} & = & (\nvarbuym,  \nvarmoneym).
 \end{array}
\]
These schemes are parametric on the propositional variables of the given language in which the various conditions and consequences are expressed. By choosing a set of values for the parameters we say that we apply the argument scheme to construct an \textbf{individual argument}. An \textbf{argument~$\Delta$,} is any (non-empty) set of individual arguments.

Let us now formalize how an argument supports a position (claim or conclusion). 
Given a cognitive state $\CalS = (\FPrem, \HPrem)$,
an individual argument $a$ \textbf{supports} $L$ iff $a=\eFact(L)$ or $a=\ehyp(L)$.
More generally, given a cognitive state $\CalS = (\FPrem, \HPrem)$,
an argument $\Delta$ \textbf{supports} $L$ iff either
\begin{enumerate}
\item there is an individual argument, $a \in \Delta$, such that
$a$ \textit{supports} $L$, or
\item there is an individual argument $a = (\{L_1, \dots, L_k\}, \Pos) \in \Delta$,
such that $L \in \Pos$ and there are $a_1, \dots, a_k \in \Delta$ such that \{$a_1, \dots, a_k$\}  
\mbox{ \textit{supports} } each one of  $L_1, \dots, L_k$.
\end{enumerate}

We will say that $\Delta$ \textbf{minimally supports} $L$ iff there is no $\Delta' \subset \Delta$ such that $\Delta'$ supports $L$.

Reasoning to a conclusion within an argumentation logic framework is based on considering arguments that support the conclusion and other related statements (e.g.\ that support the premises of a conclusion). It is important to note
that the base case of the above definition of support has as a consequence that the argumentative reasoning is \textbf{strongly grounded} on the current cognitive state of the reasoner. All arguments need to eventually be based on information that comes from the cognitive state. Then as the cognitive state changes the relevant arguments can change.

Consider again the above example with $\CalS' = (\FPrem', \HPrem') = (\{\varneedm\}, \{ \varneedm,\ \varasksm,\ \varbuym,\ \varmoneym  \})$. We can construct the argument 
\[
\begin{array}{lll}
\needDneedmilk = \{  \eFact(\varneedm), \ASpsuff{\needmilk} \},
\end{array}
\]
that supports the position that \textit{I will buy milk} ($\varbuym$). 
Note that the fact scheme 
\[\begin{array}{lll}
   \eFact(\varneedm) & = & (\emptyset, \varneedm),
  \end{array}
\]
is grounded in the above specified state $\CalS'$ because $\varneedm \in \FPrem$,
which in turn, is the premise 
of $\ASpsuff{\needmilk}$, that supports $\varbuym$.
%

\subsection{Attack and Defense between Arguments}

The last two elements of an argumentation logic framework, \mbox{$\ALF = \langle \ASset, \Cf, \St \rangle$}, are used to define the notions of attack and defense amongst arguments, on which we build the semantics of good quality or acceptable arguments then the semantics of argumentation based reasoning. 

The second element in \mbox{$\ALF = \langle \ASset, \Cf, \St \rangle$}, $\Cf$, denotes a \textbf{conflict relation} which is used to specify when arguments conflict with each other.
The conflict relation is typically based on a conflict relation defined in the language, $\CalL$, of the argumentation framework expressing which type of statements are in (direct) conflict with each other. Hence when arguments support conflicting positions then they are in conflict with each other and we say that they form a counter argument of each other.
The conflict relation can also contain elements expressing explicitly that two individual arguments are in conflict because the argument schemes that they are based on cannot be applied together.
%
%
Arguments are required to be \textbf{conflict-free}, i.e.\ they cannot support simultaneously conflicting positions, e.g. both $L$ and $\overline{L}$, or contain schemes that are explicitly in conflict.

The conflict relation defines directly the notion of attack between two arguments: an argument, $\Delta'$ \textbf{attacks} or is a \textbf{counterargument} of another argument, $\Delta$, iff there exists an $L$, such that $\Delta$ supports $L$ and $\Delta'$ supports $\overline{L}$, or they contain individual arguments whose argument schemes are in conflict.

To illustrate these notions, let us assume that we are given the current state 
\[
\CalS'' = (\FPrem'', \HPrem'') = (\{ \varneedm, \nvarmoneym \}, \{ \varneedm,\ \varasksm,\ \varbuym,\ \varmoneym  \}),
\]
 i.e.\ we are given the (factual) information, that \textit{I need milk} and \textit{I do not have enough money}.
Together with the argument, $\needDneedmilk$, that we considered above and which supports $\varbuym$, we can also construct
another argument 
\[
\begin{array}{lll}
\nmonyDmoneymilk  = \{\eFact(\nvarmoneym),  \ASpnecc{\nmoneymilk}\},
\end{array}
\]
which supports $\nvarbuym$. Note that this argument is grounded in $\CalS''$, because
$\eFact(\nvarmoneym)$ is grounded in $\FPrem$.
As $\varbuym$ and $\nvarbuym$ are in conflict with each other, these two arguments form counterarguments for each other.

The third element in \mbox{$\ALF = \langle \ASset, \Cf, \St \rangle$}, $\St$, is a binary \textbf{strength relation} among the argument schemes. This relation is required to be strongly non reflexive, i.e.\ it does not specify any argument scheme as stronger than itself. Informally, it is meant to capture the relative strength among argument schemes: Given two 
argument schemes $\AS$ and $\AS'$, 
$\AS \St \AS'$ means $\AS$ is \textbf{stronger} than $\AS'$. 
In the following, we will assume, as is typically the case, that schemes are only comparable to each other, when they are in conflict, i.e. they support opposing positions
or their schemes are in conflict. 

Recall the discussion in Section~\ref{sect:conditionals}:
(\ref{moneymilk}) blocks a possible prediction of~(\ref{needmilk}), because \textit{If I do not have enough money, then I can not buy milk} even if 
\textit{If I need milk}. We therefore assumed a cognitive principle where the associations from
necessary conditions are stronger. This is captured by:
\[
\begin{array}{lll}
\ASpnecc{\nmoneymilk} & \St & \ASpsuff{\needmilk}.
\end{array}
 \]
If this is the only strength relation defined on the individual arguments occurring in 
$\nmonyDmoneymilk$ and $\needDneedmilk$, then we will consider $\nmonyDmoneymilk$ as a stronger argument than $\needDneedmilk$, in the sense that $\nmonyDmoneymilk$ contains an individual argument that is stronger than some individual argument in $\needDneedmilk$ but not vice versa. 

Following the discussion in Section~\ref{sect:maxqual} and Section~\ref{sect:maxrel}, 
factual schemes are stronger than any other (opposing) scheme, and hypotheses schemes are weaker than any other (opposing) scheme.
For the running example, the \textit{strength of facts principle} gives us (among others),
the following relations:
\[
\begin{array}{l} 
\eFact(\buymilk) \St
\ASpnecc{\nmoneymilk}, \quad\eFact(\buymilk) \St 
\ehyp(\didntbuy), \\
 \eFact(\didntbuy) \St
\ASpsuff{\needmilk}, \hspace{1cm} \eFact(\didntbuy) \St
\ehyp(\buymilk).
\end{array}
\]
%

The notion of defense between arguments is similar to that of attack, but where now we additionally require that, informally, the defending argument is not of lower strength than the argument it is defending against. In other words, for an argument $\Delta$ to defend against $\Delta'$, 
$\Delta$ must be stronger than $\Delta'$ or at least of the same strength. 

Formally, given $\langle \ASset, \Cf, \St \rangle$, an argument $\Delta$ \textbf{defends}
against another argument~$\Delta'$ iff there exists an $L$ and $\Delta_{min} \subseteq \Delta$, $\Delta'_{min} \subseteq \Delta'$ such that
\begin{enumerate}\label{defense}
 \item\label{defense:cond1} $\Delta_{min}$, $\Delta'_{min}$ minimally support $L$, $\overline{L}$ respectively,
or \\ $\Delta_{min}$, $\Delta'_{min}$ minimally support the claim of an argument $a \in \Delta_{min}$ , $a' \in \Delta'_{min}$ respectively, such that (the argument schemes) $a$ and $a'$ are in conflict, and
 \item\label{defense:cond2} if there exists $a' \in \Delta'_{min}$, $a \in \Delta_{min} $ such that $a' \St a$ then \\
there exists $b \in \Delta_{min}$, $b' \in \Delta'_{min}$, such that $b \St b'$.
\end{enumerate}

In this definition, the first condition
simply requires that $\Delta$ is in conflict with $\Delta'$ while the second condition uses the strength relation on the individual arguments to lift this to a strength relation on sets of individual arguments.
%
The particular way that this lifting is done can vary in different argumentation frameworks. Here we adopt a specific choice of this based on a ``weakest link'' principle that gives a liberal form of defense.\footnote{This choice has been proven useful in a variety of different problems in AI that were studied under the particular GORGIAS preference based argumentation framework \cite{KMD94,Gorgias2019} on which we are more particularly basing our framework of Cognitive Argumentation.} It says that
if an argument $\Delta$ contains at least one individual argument which is stronger than an individual argument in another conflicting argument $\Delta'$ then $\Delta$ can defend against  $\Delta'$. Otherwise, $\Delta$ can only defend against $\Delta'$ if $\Delta$ does not contain any weaker individual elements, i.e.\ that the individual arguments in the two arguments are non-comparable under the strength relation.
In the milk example above, given $\CalS''$,
$\nmonyDmoneymilk$ defends against $\needDneedmilk$, but not vice versa.

\subsection{Acceptability of Arguments and Conclusions}
\label{sect:accept}

We are now ready to give the formal notion of good quality or acceptable arguments and using this to 
formulate an argumentation-based form of reasoning. 

Informally, an acceptable argument is one that can defend against all its counterarguments. 
Formally, given an argumentation logic framework,~\mbox{$\ALF = \langle \ASset, \Cf, \St \rangle$,} and 
a cognitive state $\CalS = (\FPrem, \HPrem)$, an argument $\Delta$ is \textbf{acceptable or admissible} in~\mbox{$\ALF (\Prem)$} iff
$\Delta$ is conflict-free and $\Delta$ defends against all arguments attacking $\Delta$.
A statement, $L$, is \textbf{acceptable} or a \textbf{credulous conclusion} of~$\ALF (\Prem)$, 
iff there exists an acceptable argument $\Delta$ in~$\ALF(\Prem)$ that supports $L$. 
Furthermore, $L$ is a \textbf{skeptical conclusion} of $\ALF (\Prem)$
iff $L$ is a credulous conclusion of $\ALF(\Prem)$, and~$\overline{L}$ is not a credulous conclusion of $\ALF(\Prem)$,
i.e.\ there is no acceptable argument supporting~$\overline{L}$.

Credulous and skeptical conclusions represent \textbf{plausible or possible} and \textbf{definite or certain} conclusions, respectively. The existence of an acceptable argument supporting a conclusion makes this only a possible conclusion: there is a good reason to conclude this but there could also be just as good reasons to conclude something contrary to it and thus we cannot be certain or definite in our conclusion.
Only when such possibilities of the contrary do not exist then we can have a high confidence in a definite conclusion. 

Let us consider again the cognitive state $\CalS''$
and the arguments $\nmonyDmoneymilk$ and $\needDneedmilk$:
\[
\begin{array}{l@{\hspace{0.1cm}}l@{\hspace{0.1cm}}ll@{\hspace{0.1cm}}l@{\hspace{0.1cm}}l}
\nmonyDmoneymilk & \mbox{ (minimally) supports } &  \nvarbuym \quad \quad\mbox{ and } \quad\quad
\needDneedmilk & \mbox{ (minimally) supports } & \varbuym.
\end{array}
\]
There is an individual argument in $\nmonyDmoneymilk$, namely\quad $\ASpnecc{\nmoneymilk}$, 
which is stronger than an argument in $\needDneedmilk$,
namely \quad $\ASpsuff{\needmilk}$, but not vice versa. Therefore, 
$\nmonyDmoneymilk$ defends against $\needDneedmilk$, but not vice versa.
Accordingly, $\nvarbuym$ is acceptable but $\varbuym$ is not acceptable in $\CalA_{\varmilk} (\Prem'')$.
Therefore $\nvarbuym$ is a skeptical or definite conclusion of $\CalA_{\varmilk} (\Prem'')$.

Note that technically an acceptable argument $\Delta$ for a conclusion $L$ may need to contain arguments to defend against counterarguments on the subset of $\Delta$ that minimally supports~$L$. To illustrate this, consider again $\Prem'$, which is similar to $\Prem''$, but does not contain the fact $\nvarmoneym$. Note that  $\needDneedmilk$ would not be acceptable by itself. The argument:
\[\Dnmoneymilk = \{ \hyp{\nvarmoneym}, \ASpnecc{\nmoneymilk} \},\]
forms an attack against which $\needDneedmilk$ cannot defend against by itself (because $\needDneedmilk$
does not contain any individual argument that is stronger than some individual argument
in $\Dnmoneymilk$).
However, 
\[
\needDneedmilkmoney = \needDneedmilk \cup \{\hyp{\varmoneym} \}
\]
is an acceptable argument (for $\buymilk$) because $\needDneedmilkmoney \mbox{ defends against } \Dnmoneymilk$ by defending
against the conclusion $\nvarmoneym$ of $\Dnmoneymilk$ since the individual arguments $\hyp{\varmoneym}$ and $\hyp{\nvarmoneym}$ are of equal strength under $\St$.

\subsection{Dialectic Argumentation Process} \label{sect:dialectic}

The semantics of acceptable arguments, as defined above, affords a natural and simple \textbf{dialectic argumentation process} for constructing such arguments.\footnote{This process has been studied extensively in the literature of 
argumentation in AI (see e.g. \cite{DungKT06,KT99}).} We will 
describe here, informally, this process and briefly point out its important links to cognitive argumentation. 


The basic steps are as follows:
\begin{description}
    \item[Step 1] construct a root argument supporting a conclusion of interest,
    \item[Step 2] consider a counterargument against the root argument,
    \item[Step 3] find a defense argument against the counterargument,
    \item[Step 4] check that this defense argument is not in conflict with the root argument,
    \item[Step 5] add this defense argument to the root argument, 
    \item[Repeat] from \textbf{Step 2}, i.e.\ consider another counterargument to the now extended root argument.
\end{description}

This is a form of a debate between supporting a position and trying to defeat it. Carrying out the process until there are no other counterarguments in \textbf{Step 2} that have not already being considered, clearly results in an extended root argument that is an acceptable argument supporting the conclusion of interest. 

%

Let us illustrate the dialectic argumentation process with the running example,
where the cognitive state is $\Prem'$ (as above).
The position of interest is $\varbuym$.
In \textbf{Step 1} we construct a root argument, 
\[\needDneedmilk = \{ \Fact{\varneedm}, \ASpsuff{\needmilk}\}, \mbox{ supporting } \varbuym.
\]
In \textbf{Step 2}, we check whether there are counterarguments against this argument. We can construct the following counterargument:
\[\Dnmoneymilk = \{ \hyp{\nvarmoneym}, \ASpnecc{\nmoneymilk} \},\]
which is grounded in $\Prem'$, because $\varmoneym \in \HPrem'$.
Can we find (\textbf{Step 3}) a defense
against $\Dnmoneymilk$?
Note that $\needDneedmilk$
cannot itself defend against $\Dnmoneymilk$, as shown in the end of Section~\ref{sect:accept}.
Nevertheless, a defense is given by the hypothesis argument $\hyp{\varmoneym} = (\emptyset, \varmoneym)$, 
which we can add (\textbf{Step 4} and \textbf{Step 5}) to $\needDneedmilk$.

This new extended argument is denoted as $\needDneedmilkmoney$. Returning to \textbf{Step 2} we look for other counterarguments.
Such a counterargument is given by the hypothesis argument, $\hyp{\nvarbuym}=(\emptyset, \nvarbuym)$, which is trivially defended against with the root argument, $\needDneedmilk$.
In other words, there is no need to find a different defense argument and extend further the current  the root argument in \textbf{Step 1}. 

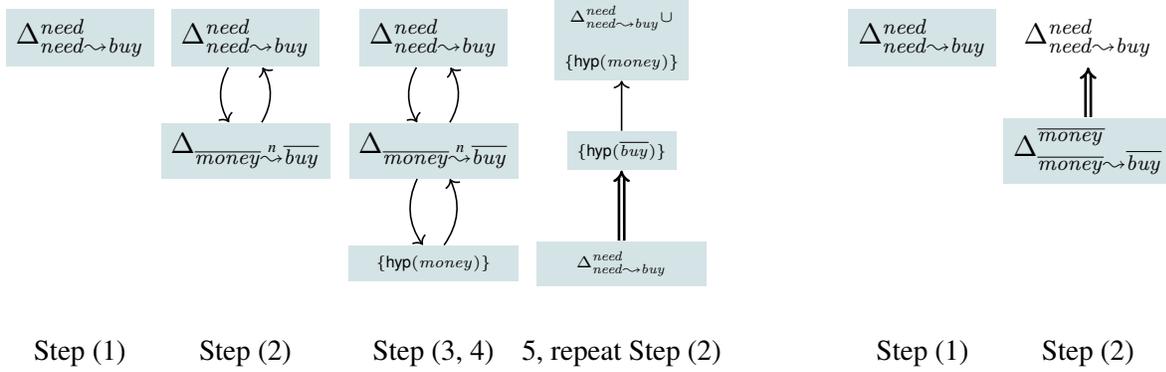
\begin{figure}
\begin{tikzpicture}[scale=0.1,-latex ,auto ,node distance =0 cm and 0.5cm ,on grid ,
semithick ,
state/.style ={ circle, minimum width =1 cm}]
\tikzstyle{l} = [draw, -latex]
\small 

\node[state,align=center,rectangle,fill=\mycolorG] (A6) {$\needDneedmilk$};

 \node (A17) [below =+4.2cm of A6] {Step (1)};

\node[state,align=center,rectangle,fill=\mycolorG] (A26) [right =+2.2cm of A6] {$\needDneedmilk$};
\node[state,align=center,rectangle,fill=\mycolorG] (A25) [below =+1.5cm of A26] {$\Dnmoneymilk$};

\draw[->] (A25) edge[bend right] node[left] {} (A26);
\draw[->] (A26) edge[bend right] node[left] {} (A25);

 \node (A27) [right =+2.2cm of A17] {Step (2)};

\node[state,align=center,rectangle,fill=\mycolorG] (A35) [right =+2.5cm of A25]  {$\Dnmoneymilk$};

\node[state,align=center,rectangle,fill=\mycolorG] (A36) [above =+1.5cm of A35] {$\needDneedmilk$};

\node[state,align=center,rectangle,fill=\mycolorG, text width=2cm] (A30) [below =1.5cm of A35] {
\tiny $\{ \thyp{\varmoneym}\}$};

\draw[->] (A35) edge[bend right] node[left] {} (A36);
\draw[->] (A36) edge[bend right] node[left] {} (A35);
\draw[->] (A30) edge[bend right] node[left] {} (A35);
\draw[->] (A35) edge[bend right] node[left] {} (A30);

 \node (A37) [below =+1.2cm of A30] {Step (3, 4)};

\node[state,align=center,rectangle,fill=\mycolorG] (A45) [right =+2.5cm of A35] {\tiny $\{ \thyp{\nvarbuym}\}$};

\node[state,align=center,rectangle,fill=\mycolorG] (A46) [above =+1.5cm of A45]  {\tiny $\needDneedmilk \cup$ \\ \tiny $\{ \thyp{\varmoneym}\}$};

\node[state,align=center,rectangle,fill=\mycolorG, text width=2cm] (A40) [below =1.5cm of A45] 
{\tiny $\needDneedmilk$};


\draw[->] (A45)-- (A46);

\draw[-Implies,line width=1pt,double distance=1pt] (A40) -- (A45);

 \node (A47) [below =+1.2cm of A40] {5, repeat Step (2)};

\node[state,align=center,rectangle,fill=\mycolorG] (A56)  [right =+4cm of A46] {$\needDneedmilk$};

 \node (A57) [below =+4.2cm of A56] {Step (1)};

\node[state,align=center,rectangle,fill=\mycolorR] (A66) [right =+2.2cm of A56] {$\needDneedmilk$};
\node[state,align=center,rectangle,fill=\mycolorG] (A65) [below =+1.5cm of A66] {$\nmonyDmoneymilk$};

\draw[-Implies,line width=1pt,double distance=1pt] (A65) -- (A66);

 \node (A67) [right =+2.2cm of A57] {Step (2)};
  
 \end{tikzpicture}

\caption{\label{fig:stepbystep:varbuym} 
Dialectic argumentation process showing that $\varbuym$ is an acceptable conclusion
given $\Prem' = (\{\varneedm\}, \{ \varneedm,\ \varasksm,\ \varbuym,\ \varmoneym  \})$
(left) and showing that $\varbuym$ is not an acceptable conclusion given
$\Prem'' = (\{\varneedm,\nvarmoneym\}, \{ \varneedm,\ \varasksm,\ \varbuym,\ \varmoneym  \})$ (right).}
\end{figure}

The dialectic argumentation process for constructing acceptable arguments can be given a tree structure. This can be illustrated by figures that show how the various arguments attack and defend each other. 
On the left of Figure~\ref{fig:stepbystep:varbuym}, the above example is shown.
On the right of Figure~\ref{fig:stepbystep:varbuym} the same process for the non-acceptability of $\needDneedmilk$ is shown given $\Prem''$, i.e.\ when $\nvarmoneym \in \FPrem''$. 
 Here and in the following figures, (temporarily) acceptable arguments are highlighted in gray and non-acceptable arguments are in white. $\uparrow$ shows attacks and/or weak defenses, i.e. defenses that are also defended against by the argument they are defending against. $\Uparrow$ shows \textit{strong} defenses, i.e. defenses that cannot be defended against by the argument they are defending against.
 In many cases these strong defenses determine the (final) acceptability of arguments.


Although the description of the dialectic argumentation process is relatively simple one can easily notice that it is computationally intensive. This is because in all the main steps, \textbf{Step 1}, \textbf{Step~2} and \textbf{Step 3} a choice is required.
What guides these choices?
In particular, within the context of Cognitive Argumentation what cognitive principles can be used to control the process?



For instance, in \textbf{Step 1}, where we construct the initial root argument, it seems natural to choose relatively strong arguments first, i.e.\ the ones based on facts rather than hypotheses. People would not start by making hypotheses, when they can choose other stronger supporting arguments based on factual information.
In \textbf{Step 2}, it is important to note that we only need to consider attacks that are strictly stronger than the root argument, avoiding to consider weaker arguments as attacks as these are already defendable by the root argument. In \textbf{Step 3}, like in \textbf{Step 1}, we would have a preference to choose strong defenses thus  minimizing the extra counterattacks that we would have to consider when we extend 
in \textbf{Step 5} the root 
argument by the defense argument. 

A thorough analysis and study of this second group of cognitive principles that are directly related to the argumentative reasoning process is outside the scope of the present paper.

%% file: bst-short.tex
We are now ready to apply the framework of Cognitive Argumentation to the suppression task. To start with we need to construct the argumentation logic framework, $\langle \ASset, \Cf, \St \rangle$, with the (argument) schemes, $\ASset$, that are relevant to the knowledge that participants have in the suppression task.

Table~\ref{tab:cogPrinSchemes} gives an overview of these schemes showing how they are motivated by the cognitive principles as introduced in the previous Sections~\ref{sect:cogprin} and~\ref{sect:argumentation}. 
%
 Note that the factual scheme is only shown for the properties $\essay$ and $\library$ as this is sufficient for the experimental setup (where the participants are only given facts on these two properties) whereas the hypothesis argument scheme can be applied on any property in the language used in the experiment. 


Motivated by the observations in Section~\ref{sect:cogprin},
the preference or strength relation, $\St$,
among these schemes in~\mbox{$\langle \ASset, \Cf, \St \rangle$}, is specified as follows:
\begin{enumerate}
\item Fact schemes are stronger than any other conflicting scheme.
\item Hypotheses schemes are weaker than any other conflicting scheme.
\item Prediction schemes established from necessary conditions ($\eASpnecc$ schemes), are stronger than conflicting schemes established from sufficient conditions ($\eASpsuff $ schemes).
\end{enumerate}

The third part of this strength relation,  as discussed before, reflects the strength of a \textit{pragmatic disabling} condition over a
\textit{motivational enabling} condition for the same consequent.

We are left to give the middle component, $\Cf$ of the realization of the cognitive argumentation framework,~\mbox{$\langle \ASset, \Cf, \St \rangle$}, for the suppression task. This conflict relation, $\Cf$, is defined through the complement relation of negation in the propositional language of the suppression task. Hence schemes (and arguments constructed from them) which support complementary positions are in conflict with each other (and hence the corresponding arguments form counterarguments of each other). In addition, we consider a second element of the conflict relation that explanation schemes with the same premises but different position can be in conflict with each other, following the general Table~\ref{tab:incompatibility} in Section~\ref{sect:conditionals}.

For example, $\ASesuff{\nlib \imply \ness}$ and 
$\ASenecc{\nlib \imply \nopen}$ are in conflict with each other as they produce different explanations for the same observation $\nlib$ (i.e.\ \textit{She did not study late in the library.})
On the other hand, $\ASesuff{\nlib \imply \ness}$  and $\ASesuff{\nlib \imply \ntbook}$
are not in conflict with each other. Rather,  both $\ness$ and $\ntbook$ form constituents parts of an explanation for $\nlib$. 
 Table~\ref{tab:incompatibility:bst}
 provides an overview of the (in)compatibility of the explanatory schemes of the suppression task, according to Table~\ref{tab:incompatibility}.
%

Finally, we note that the cognitive state, $\CalS = (\FPrem, \HPrem)$, for each group depends on
what is factually stated in the experiment and what the participants in each different group are made aware of, as presented in Section~\ref{sect:maxrel}. Hence for example, for Group I the awareness set is $\CalA = \{e, \ell \}$, whereas for Group II $\CalA = \{ e, \ell, t\}$, and for Group III, $\CalA = \{e,\ell,o\}$. These sets determine which schemes can be applied within the dialectic argumentation process.

\begin{table}\centering
\begin{tabular}{l@{\hspace{0.5cm}}l@{\hspace{0.8cm}}l@{\hspace{0.8cm}}l}
Name & Scheme & Cognitive Principle  & Sect \\\midrule
 $\Fact{e}$ &  $(\emptyset,  e  )$ & \cogPrinFact &  \ref{sect:maxqual} \\
 $\Fact{\ness}$ &  $(\emptyset,  \ness)$ & \cogPrinFact &  \ref{sect:maxqual} \\
 $\Fact{\lib}$ &  $(\emptyset,  \lib$ & \cogPrinFact &  \ref{sect:maxqual}\\
 $\Fact{\nlib}$ &  $(\emptyset,  \nlib)$ & \cogPrinFact &  \ref{sect:maxqual}\\\midrule
$\hyp{o}$ &  $(\emptyset,  o)$ & \cogPrinHyp &  \ref{sect:maxrel}\\
$\hyp{\nopen}$ &  $(\emptyset,\nopen)$ & \cogPrinHyp &  \ref{sect:maxrel}\\
$\hyp{t}$ & $(\emptyset,  t)$ & \cogPrinHyp &  \ref{sect:maxrel}\\
$\hyp{\ntbook}$ & $(\emptyset,  \ntbook)$ & \cogPrinHyp &  \ref{sect:maxrel}\\
$\hyp{e}$ & $(\emptyset,  e)$ & \cogPrinHyp &  \ref{sect:maxrel}\\
$\hyp{\ness}$ & $(\emptyset,  \ness)$ & \cogPrinHyp &  \ref{sect:maxrel}\\
$\hyp{\lib}$ & $(\emptyset,  \lib)$ & \cogPrinHyp &  \ref{sect:maxrel}\\
$\hyp{\nlib}$ & $(\emptyset,  \nlib)$ & \cogPrinHyp &  \ref{sect:maxrel} \\\midrule
$\ASpsuff{\smpl}$ & $(e,\ell)$ & \cogPrinPsuff &  \ref{sect:conditionals}.\ref{prediction} \\
$\ASpsuff{\altn}$ & $( t, \ell)$ & \cogPrinPsuff&  \ref{sect:conditionals}.\ref{prediction} \\\midrule
$\ASpnecc{\negadd}$ & $(\nopen, \nlib)$ & \cogPrinPnecc &  \ref{sect:conditionals}.\ref{prediction} \\
$\ASpnecc{\negsmpl}$ & $(\ness, \nlib)$ & \cogPrinPnecc &  \ref{sect:conditionals}.\ref{prediction}\\ \midrule
$\ASesuff{\bialt}$ & $(\ell, t)$ & \cogPrinEsuff &  \ref{sect:conditionals}.\ref{explanation}\\  
$\ASesuff{\bismpl}$ & $(\ell, e)$ & \cogPrinEsuff &  \ref{sect:conditionals}.\ref{explanation}\\ \midrule 
$\ASenecc{\negbismpl}$ & $(\nlib, \ness)$ & \cogPrinEnecc &  \ref{sect:conditionals}.\ref{explanation}\\
$\ASenecc{\negbiadd}$ & $(\nlib, \overline{o})$& \cogPrinEnecc  &  \ref{sect:conditionals}.\ref{explanation} \\ \midrule
$\ASsecpsuff{\negbismpl}$ & $(\nlib, \ness)$ & \cogPrinsecPsuff &  \ref{sect:conditionals}.\ref{secondary:prediction}\\
$\ASsecpsuff{\negbialt}$ &  $( \nlib, \ntbook)$  & \cogPrinsecPsuff&  \ref{sect:conditionals}.\ref{secondary:prediction} \\ \midrule
$\ASsecpnecc{\biadd}$ & $(\ell, o)$ &  \cogPrinsecPnecc &  \ref{sect:conditionals}.\ref{secondary:prediction}  \\
$\ASsecpnecc{\bismpl}$ & $(\ell, e)$ & \cogPrinsecPnecc&  \ref{sect:conditionals}.\ref{secondary:prediction} \\\midrule
$\ASsecesuff{\negbismpl}$ & $(\nlib, \ness)$ & \cogPrinsecEsuff &  \ref{sect:conditionals}.\ref{secondary:explanation}\\
$\ASsecesuff{\negbialt}$ &  $( \nlib, \ntbook)$ & \cogPrinsecEsuff & \ref{sect:conditionals}.\ref{secondary:explanation} \\\midrule
$\suffexo(L)$ & $(L, \exo(L))$ & \cogPrincExo &  \ref{sect:conditionals}.\ref{exognous:explanation}\\
\bottomrule
\end{tabular}
\caption{Overview of the schemes motivated by the cognitive principles introduced in Section~\ref{sect:cogprin}. \label{tab:cogPrinSchemes}}
\end{table}

\begin{table}\centering
\begin{tabular}{@{\hspace{0cm}}llG}\toprule
Observation & (In)compatibility of explanations & CL\\\midrule
 $\ell$  & $\ASesuff{\bismpl}$ and $\ASesuff{\bialt}$ are 
 incompatible & - 
\\\midrule \medskip 
$\nlib$ & $\ASsecesuff{\negbismpl}$ and $\ASsecesuff{\negbialt}$ are compatible, \\
& but incompatible with  $\ASenecc{\negbismpl}$ and $\ASenecc{\negbiadd}$ 
\\\midrule \medskip 
$\nlib$ & $\ASenecc{\negbismpl}$ and $\ASenecc{\negbiadd}$ are
incompatible, \\
& also incompatible with $\ASsecesuff{\negbismpl}$ and $\ASsecesuff{\negbialt}$\\
\midrule \medskip 
 $\ell$  & $\exo({\lib})$ is incompatible with any other explanation for $\ell$\\
 $\nlib$  & $\exo({\nlib})$ is incompatible with any other explanation for $\nlib$\\
\bottomrule
\end{tabular}
\caption{(In)compatibility of explanatory schemes depending on the observation 
and the type of condition.
\label{tab:incompatibility:bst}}
\end{table}

\subsection{Cognitive Adequacy for the Suppression Task Data} 

Participants were asked to select between three alternatives about a (natural language) statement 
$L$: (i) $L$ holds, (ii) ${L}$ does not hold and (iii) $L$ may or ${L}$ may not hold. We can see that the first two options refer to a definite conclusion for or against $L$, while the third option refers to a plausible conclusion about $L$ or its complement. It is thus important to note that the experimental conditions encourage the participants to think and reason both about definite and plausible conclusions. What is then a suitable evaluation method of a theoretical model for capturing the experimental data that evaluates the cognitive adequacy of the model?

Given that Cognitive Argumentation contains both forms of definite and plausible conclusions, this allows us to set up a criterion of evaluation of the cognitive adequacy that can probe deeply the reasoning of the approach. This criterion will be 
based on comparing the percentage of the participants' responses for a position (e.g. for \textit{She will study late in the library} ($\library$)) with the existence of acceptable arguments for the position that is asked and the existence of acceptable arguments for its complement. In particular, we will examine if in each case of the experiment: (i) there is an acceptable argument for that position asked but not for its negation (i.e.\ we have a skeptical definite conclusion),
or whether (ii) there is an acceptable argument for that position and for the negation of that position (i.e. we have a credulous plausible conclusion).

This distinction will then be required to qualitatively correspond to variations in the observed percentage of answers within the population across the three groups but also the variation within each group as  follows: 
\begin{enumerate}
\item If the population agrees on a position overwhelmingly ($> 90\%$), then
the position asked should follow in a skeptical, definite way. 
\item\label{credulous:answer} If there is no overwhelming majority, i.e.\ the percentage of the population answering for the position is less that $90\%$, then there should exist acceptable arguments for both the position and its negation, i.e.\ they both should follow credulously. 
\end{enumerate}
The second case (\ref{credulous:answer}.) reflects the fact that when there is no clear majority for the position asked, there are significant parts within the population, that believe the possibility of the opposite. In other words, although these participants may be able to build acceptable arguments for the position asked (just like those participants that have answered the question in a definite way) they also recognize the possibility of the complement of the position holding, by being able to also build an acceptable argument for the complement. 
As not the percentages of all answers were reported, but only the percentages of the answer in question, we cannot conclude that the participants who did not chose for the answer in question,
were prevented from choosing
any answer in a definite way. Even though unlikely, it is possible that these participants chose (skeptically) for the complement answer (ignoring the counterarguments).

We will now show in each of the four cases of the experiment (where different factual information is given to participants in all three groups)  how we can capture, within this framework of cognitive argumentation, the reported
experimental results. Each of the four cases will be presented in a separate subsection. In these we will introduce the main arguments that can be constructed for and against the property that is asked and show which ones are acceptable by analyzing the relevant dialectic argumentation processes. These will be illustrated by figures that show how the various arguments attack and defend each other, following the definitions introduced in Section~\ref{sect:dialectic}.

\paragraph{On Labeling Arguments, Schemes, Attacks and Acceptability}

In the sequel, we will use superscripts and subscripts in the name of an argument, $\Delta$, to indicate 
the scheme types applied for the construction of the argument: Superscripts denote \textit{fact schemes} (if applicable), whereas subscripts will denote all other schemes. A (curly) arrow used as a subscript will indicate the association captured by an argument scheme.  
Arrows are labeled by $\overset{\textit{\tiny s}}{\imply}$ referring to schemes based on a \textit{s}ufficient condition, or as $\overset{\textit{\tiny n}}{\imply}$ referring to schemes based on a \textit{n}ecessary condition.

For the figures, the interpretation of arrows and acceptability
is as in Section~\ref{sect:dialectic}: Acceptable arguments are highlighted in gray and non-acceptable arguments stay white. $\uparrow$ shows attacks and/or weak defenses, i.e.\ arguments which can also be defended against by the argument they are attacking or defending against. $\Uparrow$ shows \textit{strong} attacks and/or defenses, i.e. arguments which cannot be defended against by the argument they are attacking or defending against.

\subsection{\textit{She has an essay to finish}}\label{sub:essay}

\input{essay-short.tex}

\subsection{\textit{She does not have an essay to finish}}\label{sub:notessay}

\input{notessay-short.tex}

\subsection{\textit{She will study late in the library}}\label{sub:library}

\input{library-short.tex}

\subsection{\textit{She will not study late in the library}} \label{sub:notlibrary}

\input{notlibrary-short.tex}

\subsection{Summary of all cases of the BST experiments}

\begin{table}
 \begin{tabular}{@{\hspace{0cm}}l@{\hspace{0.2cm}}l@{\hspace{0.2cm}}cc@{\hspace{0.2cm}}ccc@{\hspace{0.2cm}}c@{\hspace{0.5cm}}} \toprule
  &  & \multicolumn{2}{c}{Predictive} 
  & \multicolumn{2}{c}{Explanatory} 
  & \multicolumn{2}{c}{Experimental Results} \\
    Given  &  Group       & suff\&necc         & suff                 & suff\&necc & suff      &\citeA{byrne:89}&\citeA{dieussaert:2000}\\
\midrule
$e$   & I      &  $\lib$             & \high{$\lib$}        & -           & -         & \high{96\% $\lib$}&\high{88\% $\lib$}
\\ 
$e$   & II     &     -               & \high{$\lib$}        & -           & -         &\high{96\% $\lib$}&\high{93\% $\lib$}
\\ 
$e$   & III    &\lesshigh{$\lib$,$\nlib$}       &\lesshigh{$\lib$, $\nlib$}       & -           & -         & \borderlesshigh{38\% $\lib$}&\borderlesshigh{60\% $\lib$}
\\
\midrule 
$\ness$& I     & \lesshigh{$\nlib$}      & \lesshigh{$\lib$,$\nlib$}       &  -          & -         & \lesshigh{46\% $\nlib$}&\lesshigh{49\% $\nlib$}
\\
$\ness$& II    & -                   &$\lib$,$\nlib$        & -           &  -        & \fbox{$\ $ 4\% $\nlib$}&\fbox{22\% $\nlib$}
\\
$\ness$& III   & \lesshigh{$\nlib$}      & \lesshigh{$\lib$,$\nlib$}       &  -          &  -        & \lesshigh{63\% $\nlib$}&\lesshigh{49\% $\nlib$}
\\
\midrule
$\lib$ & I     & \lesshigh{$e$}          & \lesshigh{$e$, $\ness$}            & \lesshigh{$e^{\; \star}$}             &\lesshigh{$e$,$\ness$}&
\lesshigh{71\% $e$}&\lesshigh{53\% $e$}
\\ 
$\lib$ & II    & -                   & $e$, $\ness$               & -           &$e$,$\ness$&\fbox{13\% $e$}&\fbox{16\% $e$}
\\ 
$\lib$ & III   & \lesshigh{$e$}          & \lesshigh{$e$, $\ness$}             & \lesshigh{$e^{\; \star}$}             &\lesshigh{$e$,$\ness$}&\lesshigh{54\% $e$} &\lesshigh{55\% $e$}
\\
\midrule
$\nlib$& I     &     \high{$\ness$}              &\high{$\ness$}        &
 \high{$\ness^{\; \star}$} \lesshigh{$e$,$\ness$}  & \high{{$\ness^{\; \star}$}} \lesshigh{$e$,$\ness$}    &\high{92\% $\ness$} & \lesshigh{69\% $\ness$}
\\ 
$\nlib$& II    &-               &\high{$\ness$}        &   -    & \high{{$\ness^{\; \star}$}} \lesshigh{$e$,$\ness$}          &\high{96\% $\ness$}& \lesshigh{69\% $\ness$}
\\ 
$\nlib$& III   &   \high{$\ness$}       &        \high{$\ness$}        &\lesshigh{$e$,$\ness$}  &\lesshigh{$e$,$\ness$}&\borderlesshigh{33\% $\ness$}&\borderlesshigh{44\% $\ness$}
\\
\bottomrule
 \end{tabular}
 \caption{Results of the suppression task modeled within cognitive argumentation.
Dark gray cells denote the conclusions that have been given by $\sim 90\%$,
and light gray cells denote the conclusions that have been given by a significant percentage (30\%-70\%).
The framed cells show the suppression effect.
The skeptical conclusions denoted by $^\star$ are the cases where an exogenous explanation is not considered.
\label{tab:allcases}}
\end{table}

Table~\ref{tab:allcases} gives a summary of the analysis of the cognitive argumentation reasoning for each of the four cases of the suppression task.
The first column shows the given fact: $e$, $\ness$, $\lib$ or $\nlib$,
respectively. 
The second column refers to the groups.
Column three and four denote the conclusions within cognitive argumentation derived when participants are assumed to be in the predictive mode of reasoning, where `suff\&necc' and `suff' mean that $e$ is understood as sufficient \textit{and} necessary or \textit{only} sufficient, respectively.
Note that, whenever only one conclusion is listed in any of these four columns, then the conclusion is skeptically entailed (definite), otherwise, it is credulously entailed (possible).
%
This applies analogously for the explanatory mode of reasoning in columns five and six.
In the four columns (columns 3 to 6), ~`-' means that this case does not seem to be cognitively plausible. 
%
Finally, columns seven and eight show the experimental results from~\cite{byrne:89} and~\cite{dieussaert:2000}, so that we can compare them with our conclusions. The framed cells in columns 7 and 8 show where the suppression effect occurs.

This table summarizes the analysis presented above and reveals the following two results by comparing the entries in columns 7 and 8 with the corresponding entries in columns 3-6:

\begin{description}

\item [Suppression:] Observed suppression coincides with the \textit{loss}, in the suppression group, of skeptical conclusions drawn in the other two groups. The conclusion at hand is exclusively skeptical or can be skeptical in the other two groups whereas in the suppression group it is predominantly credulous. It is therefore more likely for participants in the suppression group to consider the conclusion only plausible and hence avoid choosing the conclusion in their answer.

\item[Variation:] Observed significant variation in the answers in any case and any group coincides with the conclusion at hand been credulous.
Across all rows whenever the percentages in columns~7 and 8 are overwhelmingly high we only have skeptical conclusions in the corresponding columns 3 to 6 and whenever the percentages are split we have credulous conclusions in columns 3 to 6.

\end{description}


Hence the approach of Cognitive Argumentation offers a model that captures the suppression effect as well as offering an account for the qualitative difference in the degree of certainty, across the population, for the conclusions drawn.

These results stem from two important properties of Cognitive Argumentation: (i) its natural distinction between definite and plausible conclusions via the formal notions of skeptical and credulous conclusions,
(ii) its property to adapt to new and different forms of information resulting in a context-sensitive form of reasoning.

%% file: essay-short.tex
 In the first case all three groups were given the factual information, \textit{She has an essay to finish} ($e$), and were 
 asked whether \textit{She will study late in the library} ($\lib$). In Group~I, participants were assumed to be only aware of $e$ and $\lib$ as no
 other property is involved in the conditional information or question that they were given. 
Thus, the cognitive state that represents this group of participants is $\CalS = (\{ e\}, \{e, \ell \})$.

Figure~\ref{fig:stepbystep:e:lib} gives step-by-step the dialectic construction of the main (i.e. stronger and more cognitively plausible) arguments for $\lib$ and $\nlib$  in Group~I.
\footnote{In this figure and the following ones, the numbers in parentheses below the graphs refer to the steps in the dialectic argumentation process 
as specified in Section~\ref{sect:dialectic}. (1, $\dots)$ denotes for which position the argument is being constructed in step 1.}
The (strongest) argument supporting $\lib$ is given by combining the \textit{fact scheme} for $e$ together with the \textit{sufficient prediction scheme} for $\lib$ (Figure~\ref{fig:stepbystep:e:lib}.1, $\lib$):
\[\begin{array}{lll}
   \eDel & = & \{ \Fact{\essay},  \ASpsuff{\smpl} \}.
  \end{array}
 \]
It is easy to recognize this as a \textit{modus ponens} argument expressed here in an argumentation perspective. For supporting $\nlib$ the main argument is  constructed by applying the 
\textit{necessary prediction scheme} for $\nlib$ with 
the hypothesis scheme for~$\ness$ (Figure~\ref{fig:stepbystep:e:lib}.1, $\nlib$):
\[\begin{array}{lll}
   \Dnel & = & \{ \hyp{\ness},  \ASpnecc{\negsmpl} \}.
  \end{array}
 \]
These two arguments attack each other but as the figures show only $\eDel$ is able to defend against the attack via the argument $\eD =  \{ \Fact{\essay}\}$ that it contains. In fact, $\Dnel$ is immediately defeated by the stronger argument $\eD$ which attacks $\Dnel$ on the hypothesis part it contains and 
for which there is no defense. Consequently, $\lib$ is an acceptable (plausible) conclusion whereas $\nlib$ is not. 
 
Combining the two results for $\lib$ and $\nlib$ we see that $\ell$ is a skeptical conclusion: the \textit{modus ponens} argument of $\eDel$ prevails. This conforms with our criterion of evaluation, to reflect with a skeptical conclusion the overwhelming majority of responses for \textit{She will study late in the library} 
in this first group  ({\color{gray}{96}}/{{\color{gray}88\%}})\footnote{Here, and in the sequel, the first percentage refers to the results in~\cite{byrne:89}, and the second percentage
refers to the results  in~\cite{dieussaert:2000}.}.

 \begin{figure}

 \begin{center}
    \begin{tikzpicture}[scale=0.1,-latex ,auto ,node distance =1 cm and 2cm ,on grid ,
semithick ,
state/.style ={ circle, minimum width =1 cm}]
\tikzstyle{l} = [draw, -latex]
\node[state,align=center,rectangle,fill=\mycolorG] (A6) [above =+1cm of A5] {$\eDel$};

 \node (A17) [below =+1cm of A1] {1, $\lib$};

\node[state,align=center,rectangle,fill=\mycolorG] (A26) [right =+2cm of A6] {$\eDel$};
\node[state,align=center,rectangle,fill=\mycolorG] (A25) [below =+1cm of A26]{$\Dnel$};

\draw[->] (A25) -- (A26);

 \node (A27) [right =+2cm of A17] {2};

\node[state,align=center,rectangle,fill=\mycolorR] (A35) [right =+2cm of A25] {$\Dnel$};

\node[state,align=center,rectangle,fill=\mycolorG] (A36) [above =+1cm of A35] {$\eDel$};

\node[state,align=center,rectangle,fill=\mycolorG] (A30) [below =1cm of A35] {$\eD$};

\draw[->] (A35) -- (A36);
\draw[-Implies,line width=1pt,double distance=1pt] (A30) --  (A35);

 \node (A37) [right =+2cm of A27] {3-4};

%
%
%

\node[state,align=center,rectangle,fill=\mycolorG] (A45) [right =+4cm of A36] {$\Dnel$};



 \node (A47) [below =+3cm of A45] {1, $\nlib$};

\node[state,align=center,rectangle,fill=\mycolorR] (A55) [right =+2cm of A45] {$\Dnel$};

\node[state,align=center,rectangle,fill=\mycolorG] (A56) [below =+1cm of A55] {$\eD$};

\draw[-Implies,line width=1pt,double distance=1pt] (A56) --  (A55);

 \node (A57) [right =+2cm of A47] {2};

\end{tikzpicture}

\captionof{figure}{\label{fig:stepbystep:e:nlib}\label{fig:stepbystep:e:lib} 
Construction for $\lib$ (left) and for $\nlib$ (right) in Group~I given~$e$.}

\end{center}
\end{figure} 

 \begin{figure}
 \begin{center}
\begin{tikzpicture}[scale=0.1,-latex ,auto ,node distance =1 cm and 2cm ,on grid ,
semithick ,
state/.style ={ circle, minimum width =1 cm}]
\tikzstyle{l} = [draw, -latex]
\node[state,align=center,rectangle,fill=\mycolorG] (A6) [above =+2cm of A5] {$\eDel$};

 \node (A17) [below  =+0cm of A1] {1, $\lib$};

\node[state,align=center,rectangle,fill=\mycolorG] (A25) [right =+1.3cm of A6] {$\eDel$};

\node[state,align=center,rectangle,fill=\mycolorG] (A26) [below =+1cm of A25] {$\Dnol$};

\draw[->] (A26) -- (A25);

 \node (A27) [right =+1.3cm of A17] {2};

\node[state,align=center,rectangle,fill=\mycolorG] (A35) [right =+1.5cm of A25] {$\eDel$};

\node[state,align=center,rectangle,fill=\mycolorG] (A36) [below =+1cm of A35] {$\Dnol$};

\node[state,align=center,rectangle,fill=\mycolorG] (A30) [below =2cm of A35] {$\Do$};

\draw[->] (A36) --(A35);
\draw[->](A30) --  (A36);

 \node (A37) [right =+1.5cm of A27] {3-4};

%
%
%
%
%
%

 \node[state,align=center,rectangle,fill=\mycolorG] (A56) [right =+1.8cm of A35] {$\eDel \cup \Do$};

\node[state,align=center,rectangle,fill=\mycolorG] (A55) [below =+1cm of A56] {$\Dno$};

\node[state,align=center,rectangle,fill=\mycolorG] (A50) [below =+1cm of A55] {$\Do$};

\draw[->]  (A55) --(A56);
\draw[->] (A50) --(A55);

 \node (A47) [right =+1.9cm of A37] {5-2, repeat};

%
%

\node[state,align=center,rectangle,fill=\mycolorG] (A45) [right = +4cm of A56] {$\Dnol$};



 \node (A47) [below =+2.9cm of A45] {1, $\nlib$};

\node[state,align=center,rectangle,fill=\mycolorG] (A55) [right =+1.5cm of A45] {$\Dnol$};

\node[state,align=center,rectangle,fill=\mycolorG] (A56) [below =+1cm of A55] {$\eDel$};

\draw[->] (A56) -- (A55);

 \node (A57) [right =+1.5cm of A47] {2};

\node[state,align=center,rectangle,fill=\mycolorG] (A65) [right =+1.5cm of A55] {$\Dnol$};

\node[state,align=center,rectangle,fill=\mycolorG] (A66) [below =+1cm of A65] {$\eDel$};

\node[state,align=center,rectangle,fill=\mycolorG] (A67) [below =+1cm of A66] {$\Dnol$};

\draw[->] (A66) -- (A65);

\draw[->] (A67) --  (A66);

 \node (A68) [right =+1.5cm of A57] {3-4};
 
%
%
%
%
%
%
%
%

\end{tikzpicture}

\captionof{figure}{\label{fig:stepbystep:e:nlib:open}Construction for $\lib$ (left) and $\nlib$ (right) in Group~III given~$e$.}

\end{center}
\end{figure}
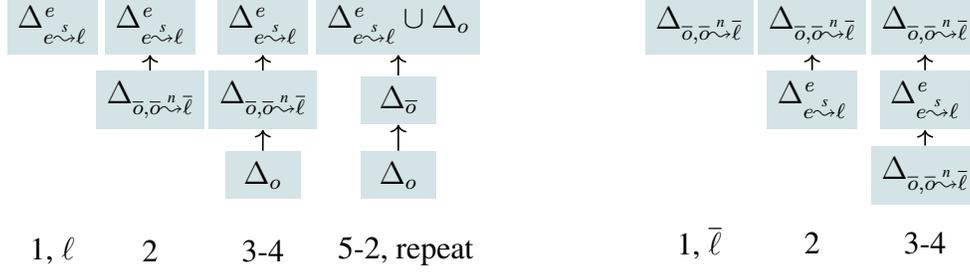

For Group~II, the argumentation analysis is essentially the same as for Group~I. The new awareness of \textit{She has a textbook to read} ($t$) does not have a significant effect. In particular, it does not introduce any new arguments 
supporting $\nlib$ and hence $\lib$ remains a skeptical conclusion, as required by the overwhelming majority
 ({\color{gray}96\%}/{\color{gray}93\%}) also in Group II. 

For Group~III, the cognitive state of its participants, is $\CalS = (\{ e\}, \{e, \ell, o \})$. Differently from Group~I and Group~II, we can now construct another strong argument for $\nlib$ based on the hypothesis prediction scheme for $\nopen$ together with 
the \textit{necessary prediction scheme} for~$\nlib$ (Figure~\ref{fig:stepbystep:e:nlib:open}, $\nlib$):
\[\begin{array}{lll}
   \Dnol & = & \{ \hyp{\nopen},  \ASpnecc{\negadd} \}.
  \end{array}
 \]
In Figure~\ref{fig:stepbystep:e:nlib:open} (left) we see how $\Dnol$ attacks the \textit{modus ponens} argument $\eDel$. This in turn is able to defend against $\Dnol$ with the help of $\Do = \{ \hyp{\open} \}$ by opposing the hypothesis part $\hyp{\nopen}$ inside $\Dnol$. Thus $\eDel \cup \Do$ can defend against all its attacks and so it is an acceptable argument for $\ell$. On the other hand, Figure~\ref{fig:stepbystep:e:nlib:open} (right) shows how $\Dnol$
supporting $\nlib$ can itself defend against its attack by 
$\eDel$, as $\ASpnecc{\negadd}$ in $\Dnol$ is stronger than $\ASpsuff{\smpl}$ in $\eDel$. Hence $\Dnol$ is an acceptable argument for $\nlib$.\footnote{The optimization of the dialectic argumentation process in Figure~\ref{fig:stepbystep:e:nlib:open} (only with strong attacks), stops after (3-5).}

Summing up, we see that in Group~III both $\lib$ and
$\nlib$, are plausible (credulous) conclusions. This then accounts for the observed suppression effect, as here only {\color{gray}38\%}/{\color{gray}60\%} concluded that \textit{She will study late in the library}. It is likely, that these participants constructed only the argument $\eDel$ for $\lib$ and hence concluded definitely $\lib$. 
It seems likely that the other {\color{gray}62\%}/{\color{gray}40\%} of the participants were able to construct both $\eDel$ and $\Dnol$ and hence were not (skeptically) sure that $\lib$ held.
However, as already mentioned earlier, the percentages of the other two answer possibilities 
were not reported, and thus possibly, some participants skeptically concluded that $\nlib$ held
(ignoring $\eDel$).

%% file: notessay-short.tex
In this second case all three groups were given the fact that \textit{She does not have an essay to finish} ($\nessay$) and were asked whether \textit{She will study late in the library}~($\library$). 
Similarly to the previous case, for Group I the corresponding cognitive state is $\CalS = (\{ \overline{e}\}, \{e, \ell \})$.

Figure~\ref{fig:stepbystep:ne} (left) shows two arguments that support $\lib$, one based on 
 the \textit{hypothesis scheme}
for $e$:
\[
  \begin{array}{lll@{\hspace{2cm}}lll}
   \Del & = & \{ \hyp{\essay},   \ASpsuff{\smpl}\},
  \end{array}
\]
and another by simply hypothesizing $\lib$, namely the argument $\Dl  =  \{ \hyp{\lib}\}$. The first argument, $\Del$, is easily defeated (i.e. attacked with no defense against the attack) by the factual argument $\neD = \{\Fact{\overline{e}} \}$ that attacks the weak part, $\hyp{\essay}$,  of $\Del$. 

The other argument, $\Dl$, for $\lib$ is defeated by a strong argument supporting $\nlib$, constructed from the \textit{factual scheme} for $\ness$ and the $\ASpnecc{\negsmpl}$ scheme:
\[
  \begin{array}{lll@{\hspace{2cm}}lll}
 \neDel & = & \{ \Fact{\overline{e}},  \ASpnecc{\negsmpl} \}.
\end{array}
\]
This is indeed a strong argument for $\nlib$ and as shown in Figure~\ref{fig:stepbystep:ne} (right) it is able to defend against its attacks. Hence 
$\nlib$ is an acceptable plausible conclusion whereas $\lib$ is not so.
Thus $\nlib$ is a skeptical definite conclusion. This corresponds to what about half of the participants ({\color{gray}{46\%}}/{\color{gray}49\%}) concluded,
namely that \textit{She will not study late in the library}.
This is similar for Group III ({\color{gray}{63\%}}/{\color{gray}49\%}).
Note that this conclusion depends on the assumption that participants consider $e$ also as a necessary condition in $\smpl$, which enables the necessary prediction scheme, $\ASpnecc{\negsmpl}$ that is in~$\neDel$.

\begin{figure}
\begin{center}

\begin{tikzpicture}[scale=0.1,-latex ,auto ,node distance =1 cm and 2cm ,on grid ,
semithick ,
state/.style ={ circle, minimum width =1 cm}]
\tikzstyle{l} = [draw, -latex]
\node[state,align=center,rectangle,fill=\mycolorG] (A6) [above =+1cm of A5] {$\Del$};

 \node (A17) [below =+1.1cm of A1] {1a, $\lib$};

\node[state,align=center,rectangle,fill=\mycolorR] (A26)  [right =+1.8cm of A6] {$\Del$};
\node[state,align=center,rectangle,fill=\mycolorG] (A25) [below =+1cm of A26]{$\neD$};

\draw[-Implies,line width=1pt,double distance=1pt] (A25) -- (A26);

 \node (A27) [right =+1.8cm of A17] {2a};

\node[state,align=center,rectangle,fill=\mycolorG] (A35) [right =+2cm of A26] {$\Dl$};




 \node (A37) [right =+2cm of A27] {1b, $\lib$};

\node[state,align=center,rectangle,fill=\mycolorR] (A45) [right =+1.5cm of A35] {$\Dl$};

\node[state,align=center,rectangle,fill=\mycolorG] (A46) [below =+1cm of A45] {$\neDnel$};


\draw[-Implies,line width=1pt,double distance=1pt]  (A46) -- (A45);

 \node (A47) [right =+1.5cm of A37] {2b};

%
%
%
%
%

\node[state,align=center,rectangle,fill=\mycolorG] (A1) {$\eDel$};

\node[state,align=center,rectangle,fill=\mycolorR] (A0) [below =1cm of A1] {$\Dnel$};

\node[state,align=center,rectangle,fill=\mycolorR] (A5) [above =+1cm of A1] {$\Dnl$};
\node[state,align=center,rectangle,fill=\mycolorG] (A6) [right =+3cm of A45] {$\neDnel$};

 \node (A17) [right =+3cm of A47] {1, $\nlib$};

\node[state,align=center,rectangle,fill=\mycolorG] (A26)  [right =+1.3cm of A6] {$\neDnel$};
\node[state,align=center,rectangle,fill=\mycolorG] (A25) [below =+1cm of A26]{$\Dl$};

\draw[->] (A25) -- (A26);

 \node (A27) [right =+1.3cm of A17] {2a};

\node[state,align=center,rectangle,fill=\mycolorR] (A35) [right =+1.3cm of A25] {$\Dl$};

\node[state,align=center,rectangle,fill=\mycolorG] (A36) [above =+1cm of A35] {$\neDnel$};

\node[state,align=center,rectangle,fill=\mycolorG] (A30) [below =1cm of A35] {$\neDnel$};

\draw[->]  (A35) edge node[left] {} (A36);
\draw[-Implies,line width=1pt,double distance=1pt] (A30) --  (A35);

 \node (A37) [right =+1.3cm of A27] {3a-4a};

\node[state,align=center,rectangle,fill=\mycolorG] (A46)  [right =+2cm of A36] {$\neDnel$};
\node[state,align=center,rectangle,fill=\mycolorG] (A45) [below =+1cm of A46]{$\Del$};

\draw[->] (A45) edge node[left] {} (A46);

 \node (A47) [right =+2cm of A37] {2b};

\node[state,align=center,rectangle,fill=\mycolorR] (A55) [right =+1.5cm of A45] {$\Del$};

\node[state,align=center,rectangle,fill=\mycolorG] (A56) [above =+1cm of A55] {$\neDnel$};

\node[state,align=center,rectangle,fill=\mycolorG] (A50) [below =1cm of A55] {$\neD$};

\draw[->]  (A55) edge node[left] {} (A56);
\draw[-Implies,line width=1pt,double distance=1pt] (A50) --  (A55);

 \node (A57) [right =+1.5cm of A47] {3b-4b};

 \end{tikzpicture}

\caption{\label{fig:stepbystep:ne} Construction for $\lib$ (left) and $\nlib$ (right) in Group~I (similarly for Group~III) given $\ness$, when $e$ is understood as sufficient \textit{and} necessary.}

\end{center}
\end{figure} 

For those participants who understand 
$e$ \textit{only} as a sufficient condition
the argumentation process is depicted in Figure~\ref{fig:stepbystep:ne:nlib:suff}. 
Figure~\ref{fig:stepbystep:ne:lib:suff} (left) shows that 
$\Dl$, can now only be attacked by the opposite hypothetical argument $\Dnl$, which it can easily defend against, and so $\lib$ is acceptable. 
\begin{figure}
\begin{center}    
\begin{tikzpicture}[scale=0.1,-latex ,auto ,node distance =1 cm and 2cm ,on grid ,
semithick ,
state/.style ={ circle, minimum width =1 cm}]
\tikzstyle{l} = [draw, -latex]

\node[state,align=center,rectangle,fill=\mycolorG] (A45) {$\Dl$};



 \node (A47) [below =+3cm of A45] {1, $\lib$};

\node[state,align=center,rectangle,fill=\mycolorG] (A55) [right =+2cm of A45] {$\Dl$};

\node[state,align=center,rectangle,fill=\mycolorG] (A56) [below =+1cm of A55] {$\Dnl$};

\draw[->] (A56) --  (A55);

 \node (A57) [right =+2cm of A47] {2};

\node[state,align=center,rectangle,fill=\mycolorG] (A65) [right =+2cm of A55] {$\Dl$};

\node[state,align=center,rectangle,fill=\mycolorG] (A66) [below =+1cm of A65] {$\Dnl$};

\node[state,align=center,rectangle,fill=\mycolorG] (A67) [below =+1cm of A66] {$\Dl$};

\draw[->] (A66) --  (A65);

\draw[->] (A67) --  (A66);

 \node (A67) [right =+2cm of A57] {3-4};

%

\node[state,align=center,rectangle,fill=\mycolorG] (A75) [right =+3cm of A65] {$\Dnl$};



 \node (A77) [below =+3cm of A75] {1, $\nlib$};

\node[state,align=center,rectangle,fill=\mycolorG] (A85) [right =+1.2cm of A75] {$\Dnl$};

\node[state,align=center,rectangle,fill=\mycolorG] (A86) [below =+1cm of A85] {$\Dl$};

\draw[->] (A86) --  (A85);

 \node (A87) [right =+1.2cm of A77] {2a};

\node[state,align=center,rectangle,fill=\mycolorG] (A95) [right =+1.2cm of A85] {$\Dnl$};

\node[state,align=center,rectangle,fill=\mycolorG] (A96) [below =+1cm of A95] {$\Dl$};

\node[state,align=center,rectangle,fill=\mycolorG] (A97) [below =+1cm of A96] {$\Dnl$};

\draw[->] (A96) --  (A95);

\draw[->] (A97) --  (A96);

 \node (A97) [right =+1.2cm of A87] {3a-4a};

\node[state,align=center,rectangle,fill=\mycolorG] (A105) [right =+2cm of A95] {$\Dnl$};

\node[state,align=center,rectangle,fill=\mycolorG] (A106) [below =+1cm of A105] {$\Del$};

\draw[-Implies,line width=1pt,double distance=1pt] (A106) --  (A105);

 \node (A107) [right =+2cm of A97] {2b};

\node[state,align=center,rectangle,fill=\mycolorG] (A115) [right =+1.5cm of A105] {$\Dnl$};

\node[state,align=center,rectangle,fill=\mycolorR] (A116) [below =+1cm of A115] {$\Del$};

\node[state,align=center,rectangle,fill=\mycolorG] (A117) [below =+1cm of A116] {$\neD$};

\draw[-Implies,line width=1pt,double distance=1pt] (A116) --  (A115);

\draw[-Implies,line width=1pt,double distance=1pt] (A117) --  (A116);

 \node (A117) [right =+1.5cm of A107] {3b-4b};

\node[state,align=center,rectangle,fill=\mycolorG] (A125) [right =+1.7cm of A115] {$\Dnl \cup \neD$};
\node[state,align=center,rectangle,fill=\mycolorR] (A126) [below =+1cm of A125] {$\Del$};

\node[state,align=center,rectangle,fill=\mycolorG] (A127) [below =+1cm of A126] {$\neD$};

\draw[-Implies,line width=1pt,double distance=1pt] (A126) --  (A125);

\draw[-Implies,line width=1pt,double distance=1pt] (A127) --  (A126);

 \node (A127) [right =+1.7cm of A117] {5b-2, repeat};

\end{tikzpicture}

\caption{\label{fig:stepbystep:ne:nlib:suff}\label{fig:stepbystep:ne:lib:suff}Construction for $\lib$ (left) and $\nlib$ (right) in Group~I (similarly for Group~II and~III) given $\ness$, when $e$ is understood \textit{only} as sufficient.}
\end{center}
\end{figure} 
Similarly, the only argument for $\nlib$ is based on the hypothesis scheme for $\nlib$.  Figure~\ref{fig:stepbystep:ne:nlib:suff} (right)
shows that $\Dnl$ is acceptable when combined with $\neD$ which is needed as a defense against the second attack of $\Del$ (Figure~\ref{fig:stepbystep:ne:nlib:suff}.2b).
Hence both $\lib$ and $\nlib$ are plausible conclusions when $e$ is understood only as a sufficient condition,
which covers the other half of participants, ({\color{gray}{54\%}}/{\color{gray}51\%}), that did not choose the definite conclusion \textit{She will not study late in the library}.

Lets us now consider Group~II where a significant suppression effect is observed. Hence, participants are additionally made aware of \textit{She might (not) have a textbook to read},
where \textit{She has a textbook to read} ($t$) is a sufficient condition for $\lib$.
The cognitive state that represents this group is $\CalS = (\{ \overline{e}\}, \{e, \ell, t \})$.
Now $e$ in $\smpl$ cannot be understood as a necessary condition anymore, 
because an alternative reason, $t$, for $\lib$ is made aware by the second conditional. Hence with the absence of $\ASpnecc{\negsmpl}$ we cannot construct a strong argument for $\nlib$.
Consequently, the majority is more likely to construct, in the way we saw above for Group I, acceptable arguments for either conclusion,~$\lib$ and~$\nlib$, and thus both follow credulously.

This accounts for the fact that the majority of the participants did not conclude a \textit{definite} answer, i.e.\ only {\color{gray}4\%}/{\color{gray}22\%} concluded that \textit{She will not study late in the library}.


%% file: library-short.tex
In this third case, all groups were 
 asked whether \textit{She has an essay to finish} ($\essay$), given the factual information that \textit{She will study late in the library} ($\lib$).
Hence the arguments that we need to consider are not for 
$\lib$ or $\nlib$ as in the previous cases but for $\essay$ and $\nessay$. 

In this case (as well as the next case of the experiment whose given factual information is \textit{She will not study late in the library}) where the factual information concerns the consequent, it is natural to assume that a significant amount of participants entered into an \textit{explanatory mode} (or diagnostic mode).
In this mode these participants tried to explain the factual observation in the context of the information that they are given in each group. 
%
We will therefore analyze both modes of reasoning, predictive and explanatory.
As above we will also consider the two cases where participants understood $e$ in the conditional statement, $\smpl$, as a sufficient and necessary
condition and others understood $e$ only as sufficient, although this will not be significant for the explanatory mode.


\begin{figure}
\begin{center}

\begin{tikzpicture}[scale=0.1,-latex ,auto ,node distance =1 cm and 2cm ,on grid ,
semithick ,
state/.style ={ circle, minimum width =1 cm}]
\tikzstyle{l} = [draw, -latex]
\node[state,align=center,rectangle,fill=\mycolorG] (A6) [above =+1cm of A5] {$\lDle$};

 \node (A17) [below =+1cm of A1] {1, $e$};

\node[state,align=center,rectangle,fill=\mycolorG] (A26)  [right =+1.3cm of A6] {$\lDle$};
\node[state,align=center,rectangle,fill=\mycolorG] (A25) [below =+1cm of A26]{$\Dcondnlne$};

\draw[->] (A25) -- (A26);

 \node (A27) [right =+1.3cm of A17] {2a};

\node[state,align=center,rectangle,fill=\mycolorR] (A35) [right =+1.3cm of A25] {$\Dcondnlne$};

\node[state,align=center,rectangle,fill=\mycolorG] (A36) [above =+1cm of A35] {$\lDle$};

\node[state,align=center,rectangle,fill=\mycolorG] (A30) [below =1cm of A35] {$\lD$};

\draw[->]  (A35) edge node[left] {} (A36);
\draw[-Implies,line width=1pt,double distance=1pt] (A30) --  (A35);

 \node (A37) [right =+1.3cm of A27] {3a-4a};

\node[state,align=center,rectangle,fill=\mycolorG] (A46)  [right =+2.5cm of A36] {$\lDle$};
\node[state,align=center,rectangle,fill=\mycolorG] (A45) [below =+1cm of A46]{$\Dne$};

\draw[->] (A45) edge node[left] {} (A46);

 \node (A47) [right =+2.5cm of A37] {2b};

\node[state,align=center,rectangle,fill=\mycolorR] (A55) [right =+1.3cm of A45] {$\Dne$};

\node[state,align=center,rectangle,fill=\mycolorG] (A56) [above =+1cm of A55] {$\lDle$};

\node[state,align=center,rectangle,fill=\mycolorG] (A50) [below =1cm of A55] {$\lDle$};

\draw[->]  (A55) edge node[left] {} (A56);
\draw[-Implies,line width=1pt,double distance=1pt] (A50) --  (A55);

 \node (A57) [right =+1.3cm of A47] {3b-4b};





\node[state,align=center,rectangle,fill=\mycolorG] (A65) [right = +3cm of A56] {$\Dne$};



 \node (A67) [below =+3cm of A65] {1a, $\ness$};

\node[state,align=center,rectangle,fill=\mycolorR] (A75) [right =+1.5cm of A65] {$\Dne$};

\node[state,align=center,rectangle,fill=\mycolorG] (A76) [below =+1cm of A75] {$\lDle$};

\draw[-Implies,line width=1pt,double distance=1pt] (A76) --  (A75);

 \node (A77) [right =+1.5cm of A67] {2a};

\node[state,align=center,rectangle,fill=\mycolorG] (A85) [right =+2.5cm of A75] {$\Dcondnlne$};

\node (A87) [right =+2.5cm of A77] {1b, $\ness$};

\node[state,align=center,rectangle,fill=\mycolorR] (A95) [right =+1.5cm of A85] {$\Dcondnlne$};

\node[state,align=center,rectangle,fill=\mycolorG] (A96) [below =+1cm of A95] {$\lD$};

\draw[-Implies,line width=1pt,double distance=1pt] (A96) --  (A95);

 \node (A97) [right =+1.5cm of A87] {2b};

\end{tikzpicture}

\caption{\label{fig:stepbystep:lib:e:nec}\label{fig:stepbystep:lib:ne:nec} Construction for $e$ (left) and $\ness$ (right) in Group~I  (similarly for Group~III) given $\lib$, when $e$ in $\smpl$ is understood as sufficient \textit{and} necessary.}
\end{center}
\end{figure}

\subsubsection{Prediction mode: \textit{She has an essay to finish} is necessary and sufficient}

For those participants in Group I who understood $e$ as sufficient and necessary in $\smpl$, 
a (strong) argument for $e$ is based on the \textit{fact scheme} for $\lib$ together with 
the \textit{secondary necessary prediction scheme} for 
$e$:
\[\begin{array}{lll}
  \lDle & = & \{ \Fact{\lib},  \secpnecc(\bismpl) \}.
  \end{array}
\]
Even though
\[\begin{array}{lll}
  \Dcondnlne & = & \{ \hyp{\nlib},  \secpsuff(\negbismpl) \}
  \end{array}
\]
attacks $\lDle$, it can be easily defeated by the (strong) factual argument, $\lD  =  \{ \Fact{\lib} \}$, as Figure~\ref{fig:stepbystep:lib:e:nec} (left) shows. Indeed, $\Dcondnlne$ is \textit{cognitively improbable}, i.e.\ it is unlikely to be considered by someone exactly because it directly contradicts the known fact that \textit{She will study late in the library}. 
Similarly, the other counterargument against $\lDle$, namely $\Dne  =  \{ \hyp{\ness} \}$ can be simply defended against by $\lDle$ itself. 

Figure~\ref{fig:stepbystep:lib:ne:nec} (right) shows that no argument for $\ness$ is acceptable.
In both cases, they have strong counterarguments ($\lDle$ and $\lD$, respectively), which they cannot defend against. 
Summing up, when $e$ in $\smpl$ is understood also as a necessary condition, in a predictive mode or reasoning we can construct acceptable arguments only for $e$ and thus $e$ follows skeptically. This captures the observation that a significant part of the participants, {\color{gray}{71\%}}/{\color{gray}{53\%}} concluded that \textit{She has an essay to finish}.

\subsubsection{Explanatory mode: \textit{She has an essay to finish} is only sufficient}

The participants in Group~I ({\color{gray}{29\%}}/{\color{gray}{47\%}}) who were not sure whether \textit{She has an essay to finish} holds, might have entered 
in explanatory mode. 
%
Thus explanatory schemes can be applied to explain the given observation, that \textit{She will study late in the library}.
 Furthermore, participants who did not understand $e$ in $\smpl$ as necessary may have considered that there is a possibility of some other unknown explanation for $\lib$, thus using the
exogenous explanation scheme. Such alternative exogenous explanation can help to support the opposite hypothesis of an explanation: in this case to support the hypothesis $\ness$ opposing the explanation $e$.

\begin{figure}
\begin{center}

\begin{tikzpicture}[scale=0.1,-latex ,auto ,node distance =1 cm and 2cm ,on grid ,
semithick ,
state/.style ={ circle, minimum width =1 cm}]
\tikzstyle{l} = [draw, -latex]
\node[state,align=center,rectangle,fill=\mycolorG] (A6) [above =+1cm of A5] {$\lDleSuff$};

 \node (A17) [below =+1.25cm of A1] {1, $e$};

\node[state,align=center,rectangle,fill=\mycolorG] (A26)  [right =+2cm of A6] {$\lDleSuff$};
\node[state,align=center,rectangle,fill=\mycolorG] (A25) [below =+1.1cm of A26]{$\lDexo$};

\draw[->] (A25) -- (A26);

 \node (A27) [right =+2cm of A17] {2};

\node[state,align=center,rectangle,fill=\mycolorG] (A36)  [right =+2cm of A26] {$\lDleSuff$};
\node[state,align=center,rectangle,fill=\mycolorG] (A35) [below =+1.1cm of A36]{$ \lDexo$};

\node[state,align=center,rectangle,fill=\mycolorG] (A34) [below =+1.1cm of A35] {$\lDleSuff$};

\draw[->] (A35) -- (A36);

\draw[->] (A34) -- (A35);

 \node (A37) [right =+2cm of A27] {3-4};

%

\node[state,align=center,rectangle,fill=\mycolorG] (A45) [right=+3cm of A36] {$\Dne$};



 \node (A47) [below =+3.25cm of A45] {1, $\ness$};

\node[state,align=center,rectangle,fill=\mycolorG] (A55) [right =+2cm of A45] {$\Dne$};
\node[state,align=center,rectangle,fill=\mycolorG] (A56) [below =+1.1cm of A55] {$\lDleSuff$};

\draw[-Implies,line width=1pt,double distance=1pt] (A56) --  (A55);

\node (A57) [right =+2cm of A47] {2};

\node[state,align=center,rectangle,fill=\mycolorG] (A65) [right =+2cm of A55] {$\Dne$};
\node[state,align=center,rectangle,fill=\mycolorG] (A66) [below =+1.1cm of A65] {$\lDleSuff$};
\node[state,align=center,rectangle,fill=\mycolorG] (A67) [below =+1.1cm of A66] {$
\lDexo$};

\draw[-Implies,line width=1pt,double distance=1pt] (A66) --  (A65);
\draw[->] (A67) --  (A66);

 \node (A67) [right =+2cm of A57] {3-4};

\end{tikzpicture}
 \caption{\label{fig:stepbystep:lib:e:suff}\label{fig:stepbystep:lib:ne:suff} Construction for $e$ (left) and $\ness$ (right) in Group~I (similarly for Group~II and~III) given $\lib$, when $e$ is understood \textit{only} as sufficient.
}


\end{center}
\end{figure}
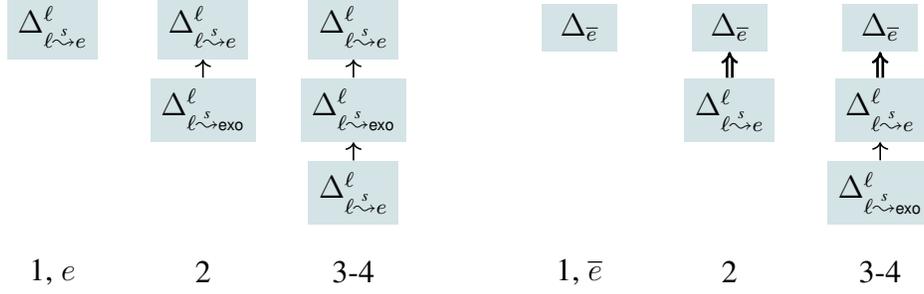

Figure~\ref{fig:stepbystep:lib:e:suff} (left) shows the acceptability of the following explanatory argument supporting $e$:
\[
\begin{array}{lll}
 \lDleSuff & =&  \{ \Fact{\lib}, \ASesuff{\bismpl} \},
\end{array}
\]
which can be attacked by
\[
\begin{array}{lll}
 \lDexo & =&  \{ \Fact{\lib}, \suffexo(\lib \imply \exo(l)) \},
\end{array}
\]
 an argument for an alternative explanation constructed via the explanation scheme, $\suffexo(l) = (\lib, \exo(l))$.
$\lDleSuff$ and $\lDexo$ can defend against each other and thus $\lDleSuff$ is acceptable.

Figure~\ref{fig:stepbystep:lib:ne:suff} (right) shows the acceptable argument supporting $\ness$.  The weak hypothesis argument $\Dne$ is defended with the help of  $\lDexo$  and hence $\Dne \cup \lDexo$ is acceptable for $\ness$. In other words, the hypothesis that \textit{She does not have an essay to finish} can stand up as valid by assuming that there was some other unknown reason for which \textit{She will study late in the library}. 


Summarizing, we can construct acceptable arguments for both, $\essay$ and for $\nessay$, and thus $\essay$ is only a plausible (credulous) conclusions. This reflects well the other significant amount of participants ({\color{gray}{29\%}}/{\color{gray}{47\%}}) who did not choose to answer 
that \textit{She has an essay to finish}. 
Furthermore, we note, that this split in the answers can be captured for one part of the participants ({\color{gray}{29\%}}/{\color{gray}{47\%}}) the possibility of an unknown exogenous explanation and for the other part ({\color{gray}{61\%}}/{\color{gray}{53\%}}) not allowing exogenous explanations. 

Lets us now consider Group II, where a suppression effect is observed. 
In this group for most participants (i) it is unlikely to consider $e$ as a necessary condition and (ii) the
possibility of an alternative explanation, such as $\exo(l)$, for the observed fact is made explicit: As $t \in \CalA$, we can apply the explanatory scheme, $\ASesuff{\bialt} = (\lib, t)$,  from which we can construct a new argument: 
\[
\begin{array}{lll}
 \lDlt & =&  \{ \Fact{\lib},\ASesuff{\bialt} \}.
\end{array}
\]

We can then construct acceptable arguments for both $e$ and $\ness$ in the same way as we have shown above by simply replacing $\lDexo$ with this explicit form of an alternative explanation $\lDlt$: the construction is completely analogous as in Figures~\ref{fig:stepbystep:lib:e:suff}.
Accordingly, $\essay$ and $\nessay$ are credulous conclusions for most participants,
which reflects well
the suppression effect in the second group, as there was no majority (only  {\color{gray}13\%}/{\color{gray}16\%}), 
that concluded that \textit{She has an essay to finish}.

Finally, for Group~III we can apply a similar analysis as in Group I to account in an analogous way for the split in the answers: about half of the participants
({\color{gray}54\%}/{\color{gray}53\%}) concluded that \textit{She has an essay to finish}.
This is because the extra information given or made aware of in Group~III, namely the (necessary) condition \textit{the library is open} does not offer any new explanatory arguments for the given observation that \textit{She will study late in the library}.

%% file: notlibrary-short.tex
In the fourth and last case of the experiment, all groups were asked whether \textit{She has an essay to finish} ($\essay$), 
given that \textit{She will not study late in the library} ($\nlib$).
%

In this case of the experiment, there is a significant
discrepancy between the results 
provided in~\cite{byrne:89} and in~\cite{dieussaert:2000}:
Whereas in Group~I and Group~II in~\cite{byrne:89}, above 90\% answered that
\textit{She does not have an essay to finish}, in~\cite{dieussaert:2000}, only 69\% gave that answer
in the same groups. 
One way of explaining this difference, is that in~\cite{byrne:89},
(almost) all of the participants in these two groups \textit{reasoned in the prediction mode}, whereas in~\cite{dieussaert:2000} some participants in Group~I and Group~II might have reasoned in the explanation mode instead.


As in the case of Section~\ref{sub:library} we will analyze both modes of reasoning, predictive and explanatory, assuming that again it is natural for some participants to reason in 
explanatory mode as the factual information given to them concerns the consequent of the conditional(s) in the general information (or the context) with which they are given to reason with.

The cognitive states for Group~I,~II and~III 
are $(\{ \nlib \}, \{e, \ell \})$, $(\{ \nlib \}, \{e, \ell, t \})$
and $(\{ \nlib \}, \{e, \ell, o \})$, respectively.


\subsubsection{Prediction Mode}

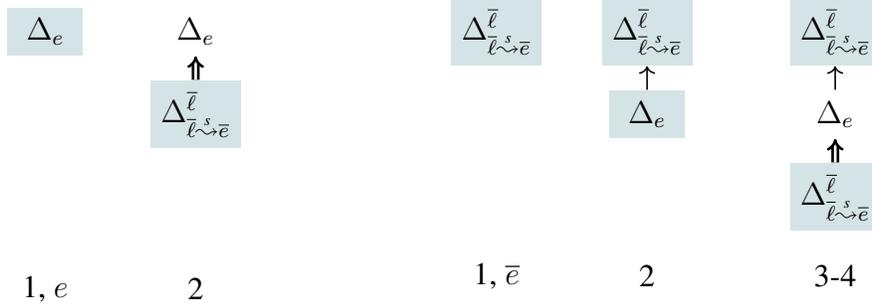
\begin{figure}
\begin{center}
\begin{tikzpicture}[scale=0.1,-latex ,auto ,node distance =1 cm and 2cm ,on grid ,
semithick ,
state/.style ={ circle, minimum width =1 cm}]
\tikzstyle{l} = [draw, -latex]

\node[state,align=center,rectangle,fill=\mycolorG] (A35) {$\De$};

 \node (A17) [below =+3.42cm of A35] {1, $e$};

\node[state,align=center,rectangle,fill=\mycolorR] (A26)  [right =+2cm of A35] {$\De$};
\node[state,align=center,rectangle,fill=\mycolorG] (A25) [below =+1.1cm of A26]{$\nlDcondnlne$};

\draw[-Implies,line width=1pt,double distance=1pt] (A25) -- (A26);

 \node (A27) [right =+2cm of A17] {2};

%
%

\node[state,align=center,rectangle,fill=\mycolorG] (A45) [right =+4cm of A26] {$\nlDcondnlne$};



 \node (A47) [below =+3.25cm of A45] {1, $\ness$};

\node[state,align=center,rectangle,fill=\mycolorG] (A55) [right =+2cm of A45] {$\nlDcondnlne$};
\node[state,align=center,rectangle,fill=\mycolorG] (A56) [below =+1.1cm of A55] {$\De$};

\draw[->] (A56) --  (A55);

\node (A57) [right =+2cm of A47] {2};

\node[state,align=center,rectangle,fill=\mycolorG] (A65) [right =+2.5cm of A55] {$\nlDcondnlne$};
\node[state,align=center,rectangle,fill=\mycolorR] (A66) [below =+1.1cm of A65] {$\De$};
\node[state,align=center,rectangle,fill=\mycolorG] (A67) [below =+1.1cm of A66] {$\nlDcondnlne$};

\draw[->] (A66) --  (A65);
\draw[-Implies,line width=1pt,double distance=1pt] (A67) --  (A66);

 \node (A67) [right =+2.5cm of A57] {3-4};

\end{tikzpicture}
 \captionof{figure}{\label{fig:stepbystep:nlib:ne}\label{fig:stepbystep:nlib:e} Construction for
$e$ (left) and $\ness$ (right) in Group~I (similarly for Group~II) given $\nlib$, 
when $e$ is understood as sufficient (prediction mode).
}
\end{center}
\end{figure}

Let us first consider those participants that reason in the \textit{prediction mode} and ask if they can build acceptable arguments for $e$ or $\ness$.
For Group I, a cognitively plausible and strong argument for $\ness$ is:  
\[
\begin{array}{lll}
\nlDcondnlne & = &  \{ \Fact{\nlib}, \secpsuff(\negbismpl) \}.
 \end{array}
\]
which is able to defend against any of its attacks 
(see Figure~\ref{fig:stepbystep:nlib:ne}, right). Hence 
$\nlDcondnlne$ is an acceptable argument for $\ness$. 
For supporting $e$, a possible argument 
consists of the hypothesis scheme for $e$ (Figure~\ref{fig:stepbystep:nlib:e}, left):
\[
\begin{array}{lll}
\De & = & \{ \hyp{e} \}.
 \end{array}
\]
However, $\De$ is attacked by $\nlDcondnlne$ 
against which $\De$ cannot defend, and so $\De$ is not acceptable. 

Note that this attack against $\De$ by $\nlDcondnlne$ reflects the informal counterargument that \textit{If she \underline{had} an essay to finish, then she \underline{would} study late in the library} but we are told that \textit{She will not study late in the library}. In more formal terms this is a \textit{Reductio ad Absurdum} counterargument rendering the hypothesis that \textit{She has an essay to finish} as non acceptable. Similarly, the argument $\nlDcondnlne$ for~$\ness$ (Figure~\ref{fig:stepbystep:nlib:ne}, right), can be recognized as reasoning with \textit{modus tollens} in an argumentation perspective.


Hence, there is no acceptable argument for $e$, and thus $\ness$ is a definite skeptical conclusion. This corresponds well with the high majority of the participants ({\color{gray}{92\%}})
in~\cite{byrne:89} who concluded and answered that \textit{She does not have an essay to finish}.

The case for Group II is analogous to that for Group I
conforming with the same observed result ({\color{gray}{96\%}}).
%
Also for Group III, participants who reason in a predictive mode will reach the same result of $\ness$ being a skeptical conclusion. But this is \textbf{not observed} in the experimental data where we have a significant reduction in the number of participants who answer that $\ness$ holds. In addition, as mentioned above, in the second experiment in~\cite{dieussaert:2000} even for Groups I and II the percentage of participants that answered that $\ness$ holds is only {\color{gray}{69\%}}, thus not all participants (skeptically) concluded $\ness$. 
One way to account for these two observations is to consider that a significant number of participants in Group III and similarly in Groups~I and~II in the \cite{dieussaert:2000} experiment, reason in an explanation mode, as we will describe below.

\subsubsection{Explanatory Mode}

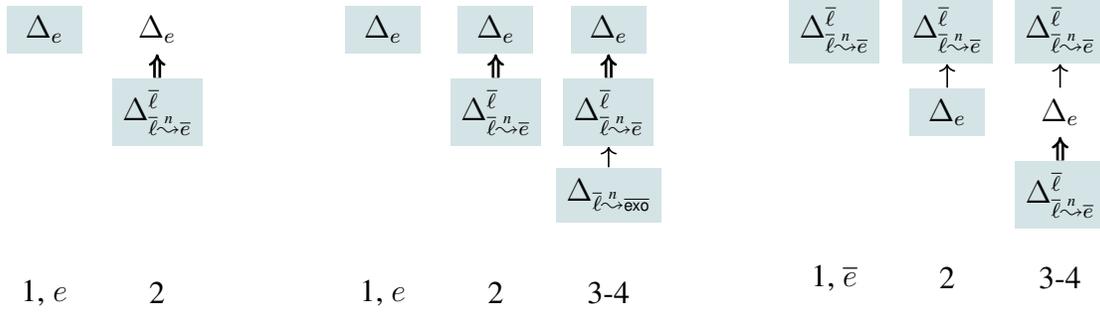
\begin{figure}
\begin{center}

\begin{tikzpicture}[scale=0.1,-latex ,auto ,node distance =1 cm and 2cm ,on grid ,
semithick ,
state/.style ={ circle, minimum width =1 cm}]
\tikzstyle{l} = [draw, -latex]
\node[state,align=center,rectangle,fill=\mycolorG] (A6) [above =+0.5cm of A5] {$\De$};

 \node (A7) [below =+2cm of A1] {1, $e$};

\node[state,align=center,rectangle,fill=\mycolorR] (A16)  [right =+1.5cm of A6] {$\De$};
\node[state,align=center,rectangle,fill=\mycolorG] (A15) [below =+1.1cm of A16]{$\nlDcondnlneN$};

\draw[-Implies,line width=1pt,double distance=1pt]  (A15) -- (A16);

 \node (A17) [right =+1.5cm of A7] {2};


\node[state,align=center,rectangle,fill=\mycolorG] (A26)  [right =+3cm of A16] {$\De$};

 \node (A27) [right =+3cm of A17] {1, $e$};

\node[state,align=center,rectangle,fill=\mycolorG] (A36)  [right =+1.5cm of A26] {$\De$};
\node[state,align=center,rectangle,fill=\mycolorG] (A35) [below =+1.1cm of A36]{$\nlDcondnlneN$};

\draw[-Implies,line width=1pt,double distance=1pt]  (A35) -- (A36);

 \node (A37) [right =+1.5cm of A27] {2};

\node[state,align=center,rectangle,fill=\mycolorG] (A46)  [right =+1.5cm of A36] {$\De$};
\node[state,align=center,rectangle,fill=\mycolorG] (A45) [below =+1.1cm of A46]{$\nlDcondnlneN$};

\node[state,align=center,rectangle,fill=\mycolorG] (A44) [below =+1.1cm of A45] {$\DcondnlnexoN$};

\draw[-Implies,line width=1pt,double distance=1pt]  (A45) -- (A46);

\draw[->] (A44) -- (A45);

 \node (A47) [right =+1.5cm of A37] {3-4};

\node[state,align=center,rectangle,fill=\mycolorG] (A55) [right =+3cm of A46]{$\nlDcondnlneN$};



 \node (A57) [below =+3.3cm of A55] {1, $\ness$};

\node[state,align=center,rectangle,fill=\mycolorG] (A65) [right =+1.5cm of A55] {$\nlDcondnlneN$};
\node[state,align=center,rectangle,fill=\mycolorG] (A66) [below =+1.1cm of A65] {$\De$};

\draw[->] (A66) --  (A65);

\node (A67) [right =+1.5cm of A57] {2};

\node[state,align=center,rectangle,fill=\mycolorG] (A75) [right =+1.5cm of A65] {$\nlDcondnlneN$};
\node[state,align=center,rectangle,fill=\mycolorR] (A76) [below =+1.1cm of A75] {$\De $};
\node[state,align=center,rectangle,fill=\mycolorG] (A77) [below =+1.1cm of A76] {$\nlDcondnlneN$};

\draw[->] (A76) --  (A75);
\draw[-Implies,line width=1pt,double distance=1pt]  (A77) --  (A76);

 \node (A77) [right =+1.5cm of A67] {3-4};

\end{tikzpicture}
 \caption{\label{fig:stepbystep:nlib:ne:necc:noexo}\label{fig:stepbystep:nlib:e:necc:noexo} \label{fig:stepbystep:nlib:e:necc}
Construction for $e$ (left), $e$ with exogenous argument (middle) and $\ness$ (right) in Group~I given $\nlib$, when $e$ is understood as sufficient and necessary (explanatory mode).
}
\end{center}
\end{figure}


In an explanatory mode of reasoning we can explain the given factual information of $\nlib$ either using the explanatory argument scheme,  $\ASenecc{\negbismpl}$, or $\ASsecesuff{\negbismpl}$, depending on whether $e$ is understood as a necessary condition for $\lib$ or not.
Accordingly, we construct the following two arguments:
\[ 
 \begin{array}{lll}
 \nlDcondnlneN= \{ \Fact{\nlib}, \ASenecc{\negbismpl} \} \quad\quad \mbox{and} \quad\quad
 \nlDcondnlneS= \{ \Fact{\nlib}, \ASsecesuff{\negbismpl} \}.
 \end{array}
\]
Figure~\ref{fig:stepbystep:nlib:ne:necc:noexo} (right) shows that 
$\nlDcondnlneN$ is acceptable. Analogously, $\nlDcondnlneS$  is also acceptable (we can replace $\nlDcondnlneN$ with $\nlDcondnlneS$).

What about arguments supporting $e$?
The hypothesis argument, $\De$ as shown in Figure~\ref{fig:stepbystep:nlib:e:necc:noexo} (left) is not acceptable. But can we find another argument to defend against its attack?
Analogously to the previous case
in Section~\ref{sub:library}, we can defend $\De$ by considering alternative explanation schemes which are incompatible with those in the attacks $\nlDcondnlneS$ and $\nlDcondnlneN$ (see Table~\ref{tab:incompatibility:bst} 
on the incompatibility between explanation schemes).
For Group I, the only other explanation is an exogenous one through which we can build an alternative explanatory argument:
 \[ 
 \begin{array}{lll}
 \DcondnlnexoN= \{ \Fact{\nlib}, \ASenecc{\nlib \imply \exo} \}.
 \end{array}
\]
Figure~\ref{fig:stepbystep:nlib:e:necc} (middle) then illustrates this 
defense through $\DcondnlnexoN$ and hence $\De \cup \DcondnlnexoN$ forms an acceptable argument for $e$. 

Summarizing the analysis for Group I, for the participants that did not have an exogenous explanation and thus $\DcondnlnexoN$ in mind, $\ness$ follows skeptically accounting thus for the experimental observation of ({\color{gray}69\%} in~\cite{dieussaert:2000}) answering that $\ness$ holds.
But when $\DcondnlnexoN$ is considered as an argument,
then both, $e$ and $\ness$ are credulous conclusions.
This accounts for the other participants in Group I , likely to be the {\color{gray}31\%} in~\cite{dieussaert:2000}, who were not sure whether $\ness$ holds.

Let us now consider Group II. In this case we have a second sufficient condition ($t$) and hence we can now construct  two arguments based on the secondary sufficient explanation for $\nlib$:
 \[ 
 \begin{array}{lll}
 \nlDcondnlneS= \{ \Fact{\nlib}, \ASsecesuff{\negbismpl} \} \quad\quad \mbox{and} \quad\quad
 \nlDcondnlntS= \{ \Fact{\nlib}, \ASsecesuff{\negbialt} \}.
 \end{array}
\]

But differently from the third case in the previous subsection, $\nlDcondnlneS$ and $\nlDcondnlntS$ are not incompatible with each other (see Table~\ref{tab:incompatibility}) but rather components of an explanation that would apply in different contexts. Hence the existence of the second argument does not affect the acceptability of the arguments for $e$ or $\ness$ 
in an explanatory mode as presented for Group~I above. 
%
This then conforms with the fact that in this case the experimental results for Group II are identical to those for Group I.

Finally, let us consider Group III. Participants were made \textit{aware of}
the possibility of an explicit or concrete alternative explanation for the given observation, namely that the reason for \textit{She will not study late in the library} is that \textit{library might not be open}. This means that participants can 
construct the argument
\[
 \begin{array}{lll}
  \nlDol = \{\Fact{\nlib}, \ASenecc{\negbiadd} \}.
 \end{array}
\]
As $\ASenecc{\negbiadd}$ is incompatible with the explanatory schemes supporting $\ness$, this new argument will defend against the arguments for $\ness$, namely against $\nlDcondnlneN$ and $\nlDcondnlneS$. Hence  $\nlDol$ can take the place of  
$\DcondnlnexoN$ in the analysis of the explanatory argumentative reasoning in the previous groups (e.g. we can replace $\DcondnlnexoN$ with $\nlDol$ in Figure~\ref{fig:stepbystep:nlib:e:necc}).
Furthermore, the acceptability of the explanatory arguments supporting $\ness$ are not affected 
by the existence of this new explanatory argument supporting $\nopen$, as these are of equal strength and therefore they can defend against each other.

Hence the suppression effect in Group III can be accounted for simply by assuming that a higher proportion of the participants (in comparison with Groups I and II) thought of an alternative explanation, i.e. other than that of $\ness$, now that they are made explicitly aware of the possible explanation of the library not being open.
Only those participants who did not consider an alternative explanation concluded for sure that $\ness$ holds and indeed the proportion of participants that did so was significantly lower ({\color{gray}{33\%}}/{\color{gray}{44\%}}).

\begin{figure}
\begin{center}


\begin{tikzpicture}[scale=0.1,-latex ,auto ,node distance =1 cm and 2cm ,on grid ,
semithick ,
state/.style ={ circle, minimum width =1 cm}]
\tikzstyle{l} = [draw, -latex]
\node[state,align=center,rectangle,fill=\mycolorG] (A6) [above =+1cm of A5] {$\De$};

 \node (A17) [below =+1.5cm of A1] {1, $e$};

\node[state,align=center,rectangle,fill=\mycolorG] (A26)  [right =+2cm of A6] {$\De$};
\node[state,align=center,rectangle,fill=\mycolorG] (A25) [below =+1.2cm of A26]{$\nlDcondnlneS$};

\draw[-Implies,line width=1pt,double distance=1pt] (A25) -- (A26);

 \node (A27) [right =+2cm of A17] {2};

 \node[state,align=center,rectangle,fill=\mycolorG] (A36)  [right =+2cm of A26] {$\De$};
\node[state,align=center,rectangle,fill=\mycolorG] (A35) [below =+1.2cm of A36]{$\nlDcondnlneS$};
\node[state,align=center,rectangle,fill=\mycolorG] (A34) [below =+1.2cm of A35]{$\nlDol$};

\draw[->] (A35) edge node[left] {} (A36);
\draw[->] (A34) -- (A35);

 \node (A37) [right =+2cm of A27] {3-4};

%

\node[state,align=center,rectangle,fill=\mycolorG] (A45) [right = +4cm of A36] {$\nlDcondnlneS$};



 \node (A47) [below =+3.25cm of A45] {1, $\ness$};

\node[state,align=center,rectangle,fill=\mycolorG] (A55) [right =+2cm of A45] {$\nlDcondnlneS$};
\node[state,align=center,rectangle,fill=\mycolorG] (A56) [below =+1.2cm of A55] {$\nlDol$};

 \draw[->] (A56) --  (A55);

\node (A57) [right =+2cm of A47] {2};

%
%

\end{tikzpicture}
 \caption{\label{fig:stepbystep:nlib:e:o:suff}\label{fig:stepbystep:nlib:ne:o:suff} Construction for $e$ (left) and $\ness$ (right) in Group~III given $\nlib$, 
when $e$ is understood \textit{only} as sufficent (explanatory mode).
%
}
\end{center}

\end{figure}
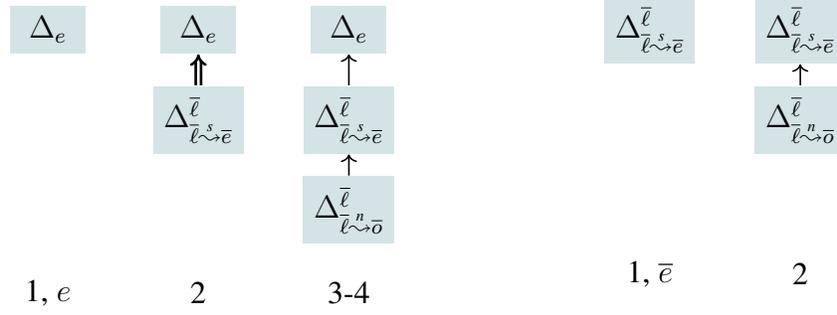

%% file: cognica.tex
Systems of Cognitive Argumentation can easily be built using existing 
argumentation technology from AI. We have developed a web based system,
called $\COGNICA$\footnote{\url{http://cognica.cs.ucy.ac.cy/}}, 
based on the underlying technology of \textit{Gorgias}~\cite{Gorgias2019}.\footnote{\url{http://www.cs.ucy.ac.cy/~nkd/gorgias/},$\ \ $ \url{http://gorgiasb.tuc.gr/}} 

The general long-term aim of $\COGNICA$ is to build a cognitive reasoner with a 
simple and natural interface so that it can be used easily by humans at large.
The purpose is to use this system (i) to evaluate the theoretical
developments in the framework of Cognitive Argumentation and (ii) to support new experimental
studies that would help with the further development of the framework.

At this initial stage of development of $\COGNICA$ our first aim is confined to testing the model of Cognitive Argumentation on human conditional reasoning and to confirm its realizability. The specific goal is to be able to reproduce the type of reasoning found in the psychological experiment of 
Byrne's suppression task and confirm the theoretical results presented in this paper. 

In designing $\COGNICA$ we have set the following functional requirements:

\begin{itemize}

\item Accept conditional common sense knowledge given in the form of controlled natural language.  

\item Reason with two levels of confidence: \textit{certain} with definite answers and
\textit{possible} with plausible answers.

\item Reason from observations to explanations.

\item Reason to conclusions based on explanations accounting for the observations.

\item Provide explanations to users on how the system has arrived at a certain definite (skeptical) or possible (credulous) conclusion.

\end{itemize}

%
%
%
%
%
%

The internal operation of the system is required to be completely transparent 
to the human user with a natural interaction which does not require the user
to have any knowledge of the underlying process of computational argumentation. 

\begin{figure}
  \includegraphics[width=\linewidth,keepaspectratio]{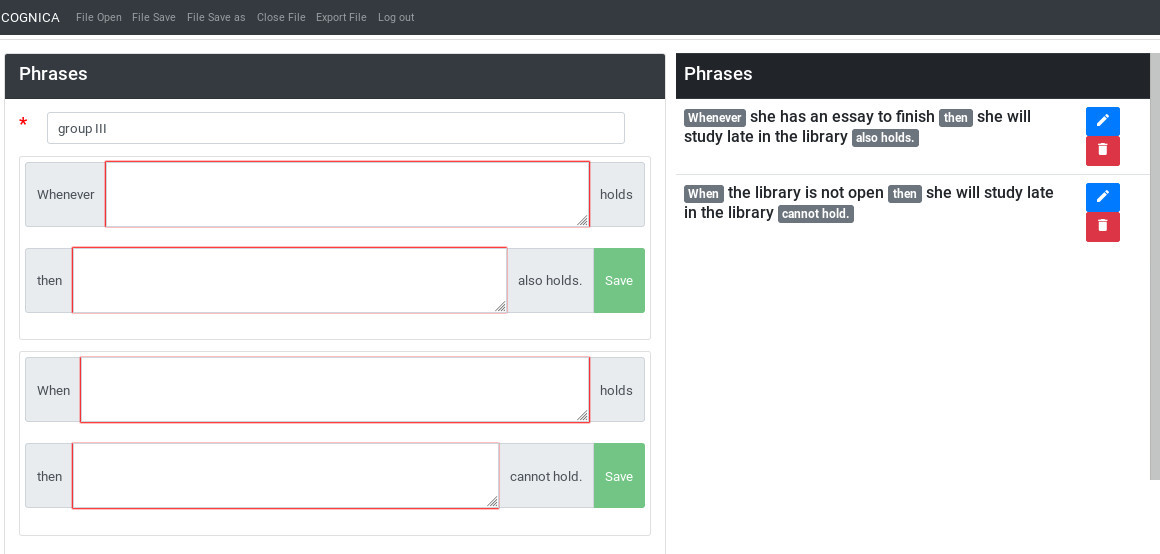}
  \caption{Knowledge for Group III as represented in \COGNICA.}
  \label{fig:GroupIII}
\end{figure}

To illustrate the system let us consider how we would use it to test 
its reasoning behaviour in the case of Byrne's suppression task. For each group 
we can enter under a different \textit{heading} or \textit{context} the general conditional information 
given to the group. Figure~\ref{fig:GroupIII} shows the case of Group III.

As can be seen from this figure, conditional knowledge that is based on 
a sufficient condition is entered in a \textit{Whenever} statement, directly linking the condition 
with the consequent, whereas knowledge based on a necessary condition 
is entered in the form of a \textit{When} statement, expressing a link between the
negation of the condition with the negation of the consequent. 

\begin{figure}
  \includegraphics[width=\linewidth,keepaspectratio]{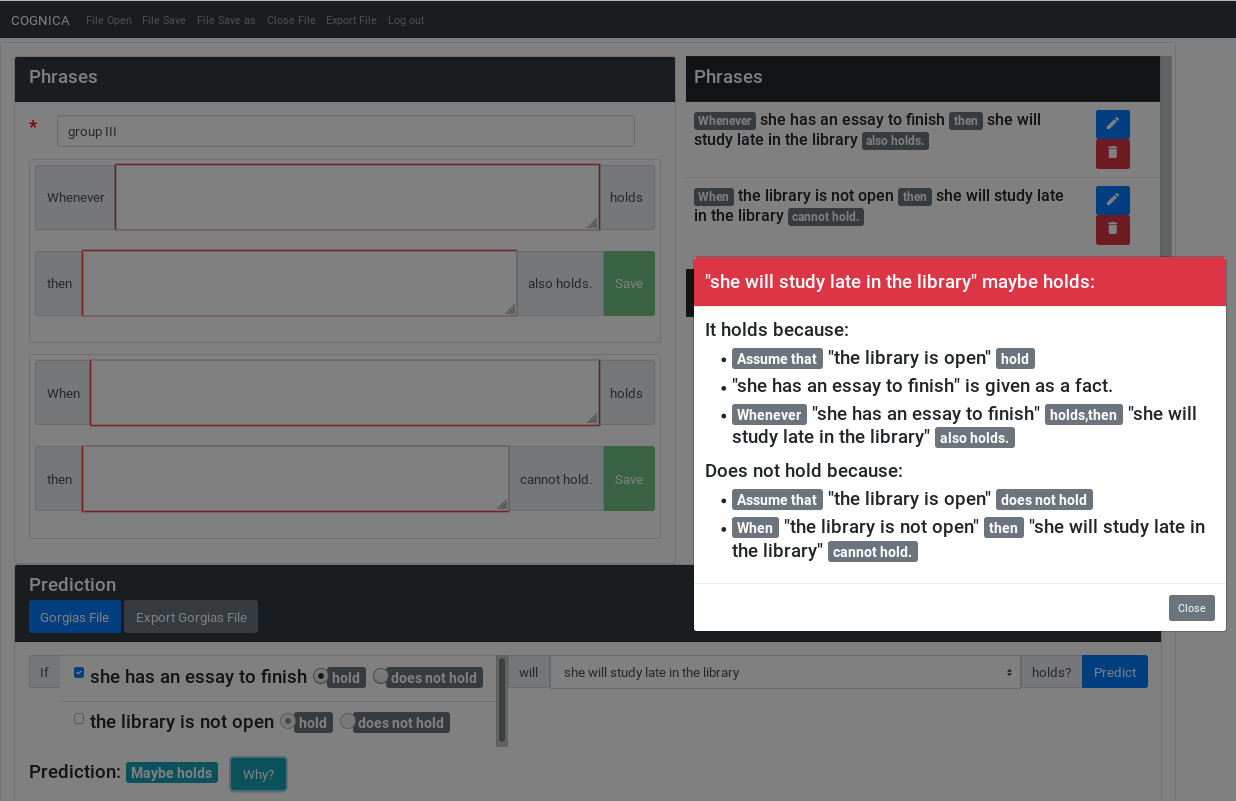}
  \caption{Case 1, \textit{She has an essay to finish}, for group~III.}
  \label{fig:GroupIIICase1}
\end{figure}

We can then select factual information and query the system for conclusions.
For example, we can select that \textit{She has an essay to finish} holds and ask whether \textit{She will study late in the library} holds. Figure~\ref{fig:GroupIIICase1}
shows this case where the answer of the system is\textit{maybe} expressing that the system considers this as possible but not definite. The figure also shows an explanation of why this is so. The explanation is presented in the form of a controlled natural language, giving the reasons for why the query may hold or may not hold, each of which reflecting the acceptable arguments in the theory 
of Cognitive Argumentation for the query and its complement (see case 1, presented in Section~\ref{sub:essay}). Similarly, we can give the system factual information asking the system to provide explanations for this and to predict if some conditions would necessarily hold or not.

As mentioned above, our future plans for developing and using the $\COGNICA$ system is to set up \textit{crowd source} experiments similar to those of Byrne's suppression task, in order to gather information on how humans reason in different circumstances. The results of these experiments will then be integrated in our plan of developing further the framework of Cognitive Argumentation.
 Another interesting question to examine is how human users are affected by the 
system in their representation and reasoning of the problem:
(i) COGNICA explicitly asks for placing conditionals within 
a \textit{when} or \textit{whenever} sentence, which forces users to consciously
distinguish between different types of conditions.
(ii) COGNICA's argumentative reasoning might affect the reasoning
of the users as soon as they are presented with the explanation of how the system arrives at its conclusions.

%% file: discussion-short.tex

We have formulated human conditional reasoning in terms of argumentation and evaluated our approach on the results of the suppression task. As Cognitive Argumentation can account for all cases of the task, it seems to form a good basis for developing a cognitively adequate model of human reasoning at large.

\subsection{Cognitive Argumentation at Large}

Let us summarize the essential elements of Cognitive Argumentation.
First, argument schemes provide a succinct form of knowledge representation,
which allows us to uniformly capture facts, hypotheses or associations between information.
Second, a strength relation between the schemes complements the representation.
It flexibly has three important properties: it is partial, qualitative and context-sensitive.
Third, reasoning is then captured via a dialectic argumentation process of attacks and defenses between arguments supporting (or not) conclusions of interest.

Argument scheme associations are not only categorized by the types of conditions they involve but also by the modes of reasoning.
Whereas \textit{modus ponens} and \textit{denial of the antecedent} belong to a predictive mode,
\textit{affirmation of the consequent} and \textit{modus tollens} belong to a explanatory (diagnostic) mode.
This classification results in important cognitive distinctions in the reasoning. For instance the discriminatory and competing
nature of explanations in explanatory reasoning is absent in predictive reasoning.
These features together help to match closer the cognitive reasoning behaviour of humans in general.
Individual differences in reasoning amount to choosing different arguments and different degrees of scrutiny apply in determining the acceptability or validity of the chosen arguments.

Another essential notion within CA is that of a cognitive state. The cognitive state represents the \textit{current context} of a limited subset of concepts and knowledge associated to them that we are (consciously) \textit{aware of} and where our attention in reasoning is currently focused.
The dialectic argumentation process of reasoning is guided by this cognitive state, which also affects when to terminate the process, even if the process has not been exhaustively carried out.
This is in accordance to human reasoning in everyday tasks, where weighing up all possibilities would be infeasible, and therefore decisions are guided by heuristically restricting our consideration.

It is important to notice that the local (conditional) knowledge in the current context is captured by argument schemes that have a simple form. We do not need to include explicitly in the premises of an argument scheme additional premises that would preclude exceptional or adversarial cases.
In other words, whenever there is no need, or \textit{awareness} made to assume differently, then the \textit{normal} case is assumed. Consider the (modus ponens) scheme $\ASpsuff{\smpl}$: the additional plethora of possible extra premises such as
\textit{She is not ill} or \textit{The deadline has not passed},
or \textit{She has not decided to drop the course},
or indeed that \textit{The library is open},
would only need to be accounted for, whenever
the current context brings them to the foreground. Whenever this is the case, 
these can be addressed through counterarguments from separate and equally simple argument schemes, as we have seen for example in the first case for Group III.

\subsection{Future Work}

The major long term challenge of CA as a cognitive model, particularly in an open setting of reasoning, lies in recognizing the context at hand within a dynamic environment, suitably adapting the argumentation framework when the context changes or is refined with further information.
What happens if Group III receives the extra information, that \textit{She has the library keys}? Will this affect their conclusions, and how? Does this only affect what argument schemes to consider, or does it also affect the strength relation between them?
We therefore need to understand in detail how does this cognitive state of humans
get formed (and altered) and how the current argumentation framework is populated with the relevant knowledge of argument schemes in the current context that we are in.

This goal drives our future work which will revolve along three aspects. 
Firstly, can we test (partially) the cognitive inferential adequacy of our CA model? If yes, how should be the 
experimental setup such that we can 
recoard how participants have arrived or are supporting their conclusions? Can we challenge humans with new counterarguments that introduce extra elements into their cognitive state of awareness and monitor their reasoning process?
Secondly, a major challenge is to link the whole framework and process of Cognitive Argumentation with Natural language, and in particular the way language is used to point to the current state of awareness and context of knowledge of argument schemes. COGNICA, is a first step towards this direction, starting from a controlled natural language user interface, which we plan to incrementally develop as close as possible to (free) natural language, employing the help of powerful existing NLP tools.
Finally, in order to understand how CA relates to practical human reasoning, we plan to apply this framework to human decision making as it is studied in the field of Behavioural Economics, based on~\cite{propecttheory:1979}.
\textit{Home economic} and similarly moral decisions taken by people at large, have been observed to deviate from logically strict but also from the \textit{rational} form of reasoning, but rather follow a heavily biased form of reasoning. As several studies have shown, small changes within the \textit{framing} of a task,
e.g.\ saying \textit{saves 90 out of 100} rather than \textit{kills 10 out of 100}, has a very significant change in the final decisions of participants~\cite{kuehberger:1995}.
Our initial investigation indicates that experimental observations of a significant change or reversal in a human decision can be accounted for in Cognitive Argumentation in a very similar manner as the suppression effect.

\subsection{Related Work}

The framework of Cognitive Argumentation is built in a multidisciplinary approach, bringing together work from AI and Cognitive Science/Psychology. In both of these areas it was long recognized that human logical reasoning would need to transcend classical formal logic. In particular, reasoning should be
non-monotonic where conclusions could not hold anymore when the context of information is changed.

The feature of non-monotonicity in logic-based AI was identified from the very start in the seminal work of McCarthy, ``Programs with Common Sense''~\cite{McCarthy:1995} and work thereafter, with a biannual workshop series starting in 1978. An intense activity on defining new non-monotonic logics followed using a variety of formal approaches.  During this however,  the original objective of modeling human commonsense reasoning shifted out of focus~\cite{McDermott87}. The proposed approaches were mostly theoretical and did not apply to real case studies, possibly because the expertise of cognitive scientists was not consulted.

An important development in the study of non-monotonic logics in AI was the relatively recent introduction of formal argumentation. Several works, see e.g.~\cite{Bondarenko97}, had shown that argumentation provides a uniform basis for reformulating most if not all different existing proposals of non-monotonic logics.
Nevertheless, the study of argumentation was also primarily motivated by and confined to the formal aspects of the framework, under the same culture that capturing human reasoning is just a matter of finding first the right logical framework: a matter of pure logic. Exceptions to this perspective, where cognitive considerations play at least a limited role, are the study of legal reasoning through argumentation~\cite{PrakkenLaw} and the more recent work on reviews and
debate analysis using argumentation technology~\cite{Reed}.

Within Cognitive Psychology/Science it was long recognized that formal logic will not serve as a good model for human reasoning. Several cognitive models have been proposed motivated primarily from experimental results of observing human reasoning and thus trying to formulate theories that would fit in the characteristics of human reasoning, such as its non-monotonic defeasible nature.
The list of these works is very long. We mention here only
some, to which our approach comes closer. These inlude the work of \cite{stenning:vanlambalgen:2008} with an approach linked with the non-monotonic nature of logic programming, the proposal of \cite{Pollock1987} where argumentation plays a significant role in a holistic proposal for cognition, similarly the approach of \cite{mercier:sperber:2011} where argumentation is considered central to human reasoning and several works on conditional reasoning whose exposition can be found in the book of \cite{nickerson:2015}. More recent work, following the Logic Programming approach of \cite{stenning:vanlambalgen:2008}, is found in \cite{declare:2017}, where cognitive principles to \textit{customize} the formal logical system was adopted and which methodology we are following in this paper.
A recent special issue of the K{\"u}nstliche Intelligenz Journal, 33(2), 2019, on Cognitive Reasoning,
exposes state of the art challenges for understanding and automating human reasoning.

Our approach concurs closely with the framework of mental models~\cite{johnsonlaird:1983,Khemlani2018FactsAP} with its emphasis on human reasoning as cases of possibilities. Argumentation is by nature a process of considering alternatives and thus argumentative reasoning is inherently possibilistic. Building acceptable arguments in Cognitive Argumentation can be seen to correspond closely to the process of constructing mental (cognitive) models.
Our correspondence of the different types of conditions, sufficient or necessary, into distinct argument schemes with a separate identity and the distinction between predictive and explanatory schemes (again drawn from the same conditional) can be reflected into the different possibilities of mental models associated to conditionals. Having drawn this parallel, it must be noted that the theory of mental models includes extensive studies on how to recognize these
distinctions from the form of the expression of a conditional in natural language and other pragmatics in the context of reasoning. This is an aspect that is currently missing in our approach and which can benefit greatly from the existing work.


On another level of comparison, as we have seen, Cognitive Argumentation supports two types of conclusions with a qualitative difference in the confidence or certainty of the conclusion: plausible or possible and definite conclusions. This stems form the relative strength between different argument schemes that is qualitative, although it can be learned through a quantitative statistical analysis from past experiences \cite{LoizosMichael2016}. As in the more recent work \cite{Khemlani2018FactsAP,Khemlanilaird:2019} within the framework of mental models, such natural distinctions between the ``degree'' of conclusions would correspond closer to real-life human reasoning than formal logical modalities of ``possible'' and ``necessary'' in the framework of Modal Logics.

%% file: conclusions.tex
We have seen how human reasoning can be formulated within a framework of dialectic argumentation,
called Cognitive Argumentation, where reasoning to conclusions is understood
as a process of contemplating between alternatives and the arguments that support them. The framework of Cognitive
Argumentation is based on the theory of computational argumentation from relatively
recent studies of argumentation in AI. It uses a variety of general cognitive principles
of reasoning, identified in Cognitive Science and Philosophy over many decades, to ``calibrate'' the
abstract and general framework from AI in order to adapt and apply it to the case
of informal human reasoning. The salient features of context-sensitive reasoning,
variability of reasoning within the population, distinguishing definite and possible conclusions, and the defeasibility of reasoning, all
emerge naturally within the framework of Cognitive Argumentation.

Cognitive Argumentation has been validated as a good cognitive adequate model of human reasoning by explaining well the
reasoning behaviors of participants in the celebrated experiments of the suppression task.
Although we have concentrated here on reasoning with conditionals the framework
is general enough to accommodate wider forms of human reasoning by suitably extending it with argument schemes appropriate for new reasoning forms and sculpting this with
further relevant cognitive principles. This is a difficult and challenging task, but one that promises to be
instructive in forming a better and more complete understanding of the relationship of
argumentation with human reasoning at large.